\documentclass[sigconf]{acmart}

\usepackage{balance}
\AtBeginDocument{%
  \providecommand\BibTeX{{%
    \normalfont B\kern-0.5em{\scshape i\kern-0.25em b}\kern-0.8em\TeX}}}

\copyrightyear{2020}
\acmYear{2020}
\setcopyright{acmcopyright}\acmConference[MM '20]{Proceedings of the 28th ACM International Conference on Multimedia}{October 12--16, 2020}{Seattle, WA, USA}
\acmBooktitle{Proceedings of the 28th ACM International Conference on Multimedia (MM '20), October 12--16, 2020, Seattle, WA, USA}
\acmPrice{15.00}
\acmDOI{10.1145/3394171.3413593}
\acmISBN{978-1-4503-7988-5/20/10}

\usepackage{bm}
\usepackage{multirow}
\begin{document}
\fancyhead{}
\title{Context-aware Feature Generation for Zero-shot Semantic Segmentation}


\author{Zhangxuan Gu}
\orcid{0000-0002-2102-2693}
\affiliation{\institution{\mbox{MoE Key Lab of Artificial Intelligence,} Shanghai Jiao Tong University}}
\email{zhangxgu@126.com}
\author{Siyuan Zhou}
\affiliation{\institution{\mbox{MoE Key Lab of Artificial Intelligence,} Shanghai Jiao Tong University}}
\email{ssluvble@sjtu.edu.cn}
\author{Li Niu*}
\affiliation{\institution{\mbox{MoE Key Lab of Artificial Intelligence,} Shanghai Jiao Tong University}}
\email{ustcnewly@sjtu.edu.cn}
\author{Zihan Zhao}
\affiliation{\institution{\mbox{MoE Key Lab of Artificial Intelligence,} Shanghai Jiao Tong University}}
\email{john745111625@gmail.com}
\author{Liqing Zhang*}
\affiliation{\institution{\mbox{MoE Key Lab of Artificial Intelligence,} Shanghai Jiao Tong University}}
\email{zhang-lq@cs.sjtu.edu.cn}
\thanks{*Corresponding authors.}

\renewcommand{\shortauthors}{Gu, et al.}
%


\begin{abstract}
Existing semantic segmentation models heavily rely on dense pixel-wise annotations. To reduce the annotation pressure, we focus on a challenging task named zero-shot semantic segmentation, which aims to segment unseen objects with zero annotations. This task can be accomplished by transferring knowledge across categories via semantic word embeddings. In this paper, we propose a novel context-aware feature generation method for zero-shot segmentation named CaGNet. In particular, with the observation that a pixel-wise feature highly depends on its contextual information, we insert a contextual module in a segmentation network to capture the pixel-wise contextual information, which guides the process of generating more diverse and context-aware features from semantic word embeddings. Our method achieves state-of-the-art results on three benchmark datasets for zero-shot segmentation. \emph{Codes are available at: https://github.com/bcmi/CaGNet-Zero-Shot-Semantic-Segmentation}
\end{abstract}

\ccsdesc[500]{Computing methodologies~Image segmentation}

\keywords{zero-shot semantic segmentation, contextual information, feature generation}


\maketitle

\section{Introduction}\label{intro}

Semantic segmentation, aiming at classifying each pixel in one image, heavily relies on the dense pixel-wise annotations~\cite{long2015fully,zhao2017pyramid,chen2018deeplab,ronneberger2015u,Lin2016RefineNet,GaoLearning}. To reduce the annotation pressure, leveraging weak annotations like image-level~\cite{oh2017exploiting,papandreou2015weakly,Yao2015Semantic}, box-level~\cite{khoreva2017simple,GraphNet}, or scribble-level~\cite{lin2016scribblesup} annotations for semantic segmentation recently gained the interest of researchers. In this work, we focus on a more challenging task named zero-shot semantic segmentation~\cite{bucher2019zero}, which further relieves the burden of human annotation. Similar to zero-shot learning~\cite{lampert2013attribute}, we divide all categories into seen and unseen categories. The training images only have pixel-wise annotation for seen categories, while both seen and unseen objects may appear in test images. Thus, we need to bridge the gap between seen and unseen categories via category-level semantic information, enabling the model to segment unseen objects in the testing stage.

Transferring knowledge from seen categories to unseen categories is not a new idea and has been actively studied by zero-shot learning (ZSL)~\cite{lampert2013attribute,akata2015evaluation,xian2019f,DBLP:conf/ijcai/GuoDHSLD19}. Most ZSL methods tend to learn the mapping between visual features and semantic word embeddings or synthesize visual features for unseen categories.

To the best of our knowledge, there are quite few works on zero-shot semantic segmentation~\cite{zhao2017open,kato2019zero,xian2019semantic,bucher2019zero}, in which only SPNet~\cite{xian2019semantic} and ZS3Net~\cite{bucher2019zero} can segment an image with multiple categories. SPNet extends a segmentation network by projecting visual features to semantic word embeddings. Since the training images only contain labeled pixels of seen categories, the prediction will be biased towards seen categories in the testing stage. Hence, they deduct the prediction scores of seen categories by a calibration factor during testing. However, the bias issue is still severe after using calibration. Inspired by feature generation methods for zero-shot classification~\cite{xian2018feature,FelixMulti}, ZS3Net learns to generate pixel-wise features from semantic word embeddings. The generator is trained with seen categories and able to produce features for unseen categories, which are then used to finetune the last $1\times 1$ convolutional (conv) layer in the segmentation network. Moreover, they extend ZS3Net to ZS3Net (GC) by using Graph Convolutional Network (GCN)~\cite{kipf2016semi-supervised} to capture spatial relationships among different objects.
However, it still has two drawbacks: 1) ZS3Net simply appends a random noise to one semantic word embedding to generate diverse features. However, the generator often ignores the random noise and can only produce limited diversity for each category-level semantic word embedding, known as mode collapse problem~\cite{xian2018texturegan,zhu2017unpaired};
2) Although ZS3Net (GC) utilizes relational graphs to encode spatial object arrangement, the contextual cues it considers are object-level and only limited to spatial object arrangement. Moreover, the relational graphs containing unseen categories are usually inaccessible when generating unseen features.

\begin{figure}[tbp]
\centering
\includegraphics[width=0.9\linewidth]{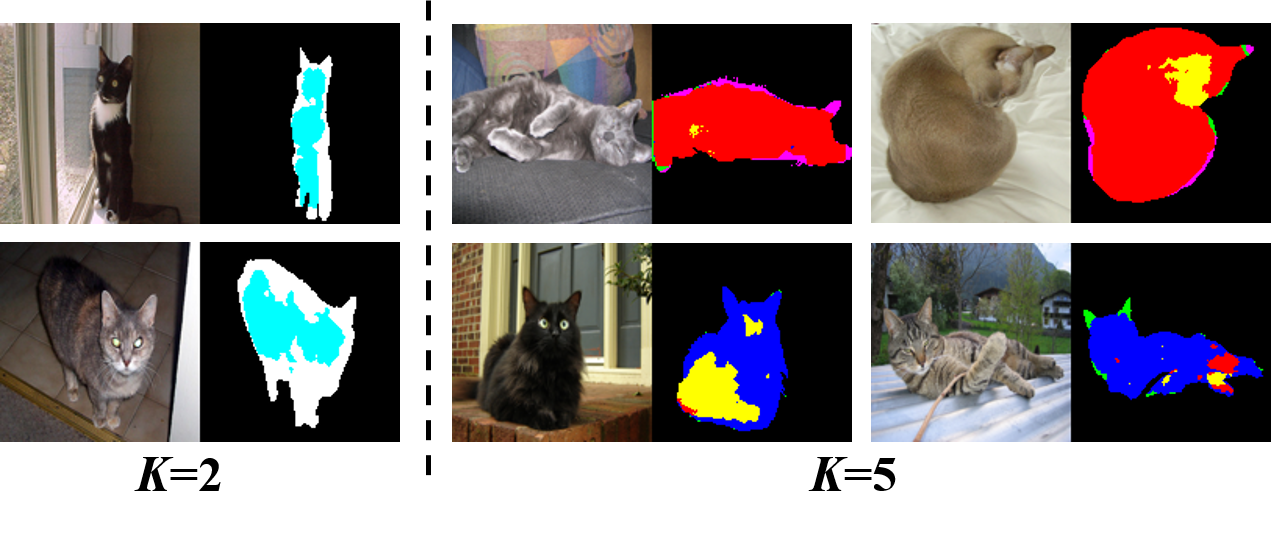}
\caption{The pixel-wise visual features of category ``cat" are grouped into $K$ clusters with each color representing one cluster in Pascal-Context. The left (\emph{resp.}, right) subfigure shows the visualization results when $K=2$ (\emph{resp.}, $K=5$).}
\label{example}
\end{figure}

In this paper, we follow the research line of feature generation for zero-shot segmentation and propose a \emph{\textbf{C}}ontextual-\emph{\textbf{a}}ware feature \emph{\textbf{G}}eneration model, CaGNet, 
by considering pixel-wise contextual information when generating features.

The contextual information of a pixel means the information inferred from its surrounding pixels (\emph{e.g.}, its location in the object, the posture of the object it belongs to,  background objects), which is not limited to spatial object arrangement considered in \cite{bucher2019zero}.
Intuitively, the pixel-wise feature vectors in deep layers highly depend on their contextual information.
To corroborate this point, we obtain the output features of the ASPP module in Deeplabv2~\cite{chen2018deeplab} for category ``cat" on Pascal-Context~\cite{mottaghi2014role}, and group those pixel-wise features into $K$ clusters by K-means. Based on Figure~\ref{example}, we observe that pixel-wise features are affected by their contextual information in an interlaced and complicated way. When $K=2$, the features from the interior (\emph{resp.}, exterior) of the cat are grouped together. When $K=5$, we provide examples in which pixel-wise features are affected by adjacent or distant background objects. For example, the red (\emph{resp.}, blue) cluster is likely to be influenced by the cushion (\emph{resp.}, green plant) as shown in the top (\emph{resp.}, bottom) row. These observations motivate us to generate context-aware features with the guidance of pixel-wise contextual information.

Unlike the feature generator in ZS3Net, which takes semantic word embedding and random noise as input to generate pixel-wise fake feature, we feed semantic word embedding and pixel-wise contextual latent code into our generator. The contextual latent code is obtained from our proposed Contextual Module ($CM$). Our $ CM $ takes the output of the segmentation backbone as input and outputs pixel-wise real feature and corresponding pixel-wise contextual latent code for all pixels. In our $CM$, we also design a context selector to adaptively weight different scales of contextual information for different pixels.
Since adequate contextual information is passed to the generator to resolve the ambiguity of feature generation, we expect that the pixel-wise contextual latent code together with semantic word embedding is able to reconstruct the pixel-wise real feature.
In other words, we build the one-to-one correspondence (bijection) between input pixel-wise contextual latent code and output pixel-wise feature. It has been proved in \cite{NeurIPS2017_6650} that the bijection between input latent code and output could mitigate the mode collapse problem, so our model can generate more diverse features from one semantic word embedding by varying the contextual latent code. We enforce the contextual latent code to follow unit Gaussian distribution to get various contextual latent codes via randomly sampling.
Therefore, the segmentation network and our feature generation network are linked by contextual module and classifier.

In summary, compared with ZS3Net, CaGNet can produce more diverse and context-aware features.
Compared with its extension ZS3Net (GC), our method has two advantages: 1) we leverage more informative pixel-wise contextual information instead of object-level contextual information;
2) we encode contextual information into latent code, which supports stochastic sampling, so we do not require explicit contextual information of unseen categories (\emph{e.g.}, relational graph) when generating unseen features. Our main contributions are:
\begin{itemize}
  \item We design a feature generator guided by pixel-wise contextual information, to obtain diverse and context-aware features for zero-shot semantic segmentation.
  \item Two minor contributions are: 1) unification of segmentation network and feature generation network; 2) contextual module with a novel context selector.
  \item Extensive experiments on Pascal-Context, COCO-stuff, and Pascal-VOC demonstrate the effectiveness of our method.
\end{itemize}

\begin{figure*}[htbp]
\centering
\includegraphics[width=0.9\linewidth]{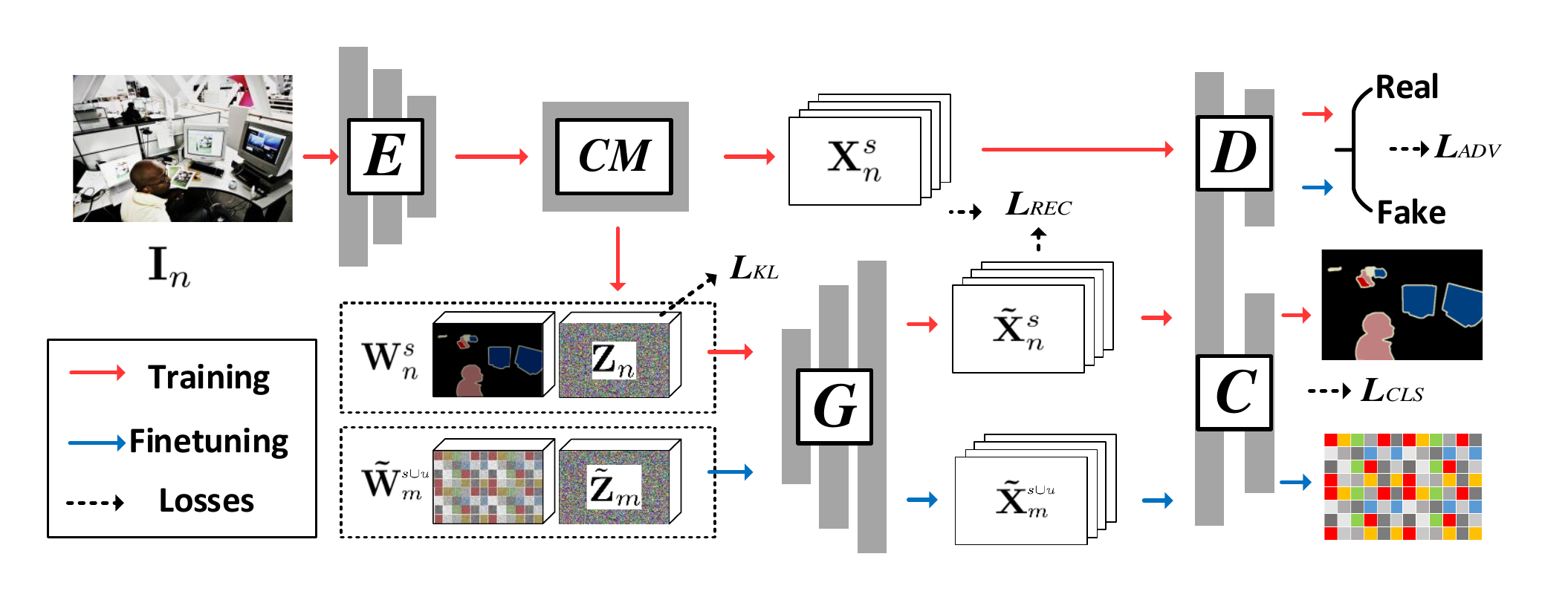}
\caption{\textbf{Overview of our CaGNet.} Our model contains segmentation backbone $E$, Contextual Module $CM$, feature generator $G$, discriminator $D$, and classifier $C$. $\textbf{W}$, $\textbf{Z}$, and $\textbf{X}$ represent semantic word embedding map, contextual latent code map, and feature map respectively (see Section~\ref{sec:CM} and \ref{sec:context_feature_generator} for detailed definition). Optimization steps are separated into training step and finetuning step indicated by two different colors (see Section~\ref{sec:optimization}).} 
\label{ours}
\end{figure*}

\section{Related Works}

\noindent\textbf{Semantic Segmentation: }State-of-the-art semantic segmentation models~\cite{long2015fully,zhao2017pyramid,chen2018deeplab,ronneberger2015u,Lin2016RefineNet,GaoLearning} are typically extending Fully Convolutional Network (FCN) \cite{long2015fully} framework with larger receptive field and more efficient encoder-decoder structure. Based on the idea to expand receptive field, PSPNet\cite{zhao2017pyramid} and Deeplab\cite{chen2018deeplab} design specialized pooling layers for fusing the contextual information from feature maps of different scales. Other methods like U-Net~\cite{ronneberger2015u} and RefineNet~\cite{Lin2016RefineNet} focus on designing more efficient network architectures to better combine low-level and high-level features.

One important characteristic of semantic segmentation is the usage of contextual information since the category predictions of target objects are often influenced by nearby objects or background scenes. Thus, many works~\cite{chen2018deeplab,Yu2015Multi} tend to explore contexts of different receptive fields with dilated convolutional layers, which also motivates us to incorporate contexts into feature generation.
However, those models still require annotations of all categories during training, and thus cannot be applied to the zero-shot segmentation task. In contrast, we successfully combine segmentation network with feature generator for zero-shot semantic segmentation.


\noindent\textbf{Zero-shot Learning: }Zero-shot learning (ZSL) was first introduced by~\cite{Lampert2009Learning}, in which training data are from seen categories, but test data may come from unseen categories. Knowledge is transferred from seen categories to unseen via category-level semantic embeddings. Many ZSL methods~\cite{akata2015label,frome2013devise,romera2015embarrassingly,fu2015transductive,10.1145/2964284.2964319,10.1145/2578726.2578746,10.1145/3240508.3240715,niu2018webly,niu2019zero-shot} attempted to learn a mapping between feature space and semantic embedding space.

Recently, a popular approach of zero-shot classification is generating synthesized features for unseen categories. For example, the method in~\cite{xian2018feature} first generated features using word embeddings and random vectors, which was further improved by later works \cite{FelixMulti,xian2019f,Mert2019Gradient,Li2019Leveraging,Mandal2019Out}. These zero-shot classification methods generated image features without involving contextual information. In contrast, due to the uniqueness of semantic segmentation, we utilize pixel-wise contextual information to generate pixel-wise features.

\noindent\textbf{Zero-shot Semantic Segmentation: }The term zero-shot semantic segmentation appeared in prior works~\cite{zhao2017open,kato2019zero,xian2019semantic,bucher2019zero}, in which only SPNet~\cite{xian2019semantic} and ZS3Net~\cite{bucher2019zero} focused on multi-category semantic segmentation. SPNet achieves knowledge transfer between seen and unseen categories via semantic projection layer and calibration method, while ZS3Net aims to generate pixel-wise features to finetune the classifier, which is biased to the seen categories. Our method is inspired by ZS3Net, but different from their method in mainly two ways: 1) we unify the segmentation network and feature generator;
2) we leverage pixel-wise contextual information to guide feature generation.
\section{Methodology}\label{method}

For ease of representation, we denote the set of seen (\emph{resp.}, unseen) categories as $\mathcal{C}^s$ (\emph{resp.}, $\mathcal{C}^u$) and $\mathcal{C}^s\cap \mathcal{C}^u=\emptyset$. In the zero-shot segmentation task, the training set only contains pixel-wise annotations of $\mathcal{C}^s$, while the trained model is supposed to segment objects of $\mathcal{C}^s\cup \mathcal{C}^u$ at test time. As mentioned in Section~\ref{intro}, the bridge between seen and unseen categories is the category-level semantic word embeddings $\{{\bar{\textbf{w}}}_c|c\in \mathcal{C}^s\cup \mathcal{C}^u\}$, in which ${\bar{\textbf{w}}}_c \in \mathcal{R}^{d}$ is the semantic word embedding of category $c$.

\subsection{Overview}

Our method, CaGNet, can be applied to an arbitrary segmentation network. We start from Deeplabv2~\cite{chen2018deeplab}, which has shown remarkable performance in semantic segmentation. Any segmentation network like Deeplabv2 can be separated into two parts: backbone $E$ and classifier $C$ (\emph{e.g.}, one or two $1\times1$ conv layers). Given an input image, the backbone outputs its real feature map, which is passed to the classifier to get the segmentation results.

To enable the segmentation network to segment unseen objects, we aim to learn a generator $G$ to generate features for unseen categories. As shown in Figure~\ref{ours}, $G$ takes the semantic word embedding map and the latent code map as input to output fake features. Then, discriminator $ D $ and classifier $ C $, with a shared $1\times 1$ conv layer, take real/fake features to output discrimination and segmentation results respectively. Note that our classifier $C$ is shared by the feature generation network and the segmentation network.
To help the generator $G$ produce more diverse and context-aware features, we insert a Contextual Module ($CM$) after the backbone $E$ of segmentation network to obtain contextual information, which is encoded into the latent code as the guidance of $G$. Therefore, we unify the segmentation network $\{E, CM, C\}$ and the feature generation network $\{CM, G, D, C\}$, which are linked by Contextual Module $CM$ and classifier $C$. Next, we will detail our $CM$ in Section~\ref{sec:CM} and feature generator in Section~\ref{sec:context_feature_generator}.
For ease of description, we use capital letter in bold (\emph{e.g.}, $\textbf{X}$) to denote a map and small letter in bold (\emph{e.g.}, $\textbf{x}_i$) to denote its pixel-wise vector. We use upperscript $s$ (\emph{resp.}, $u$) to indicate seen (\emph{resp.}, unseen) categories.

\subsection{Contextual Module}\label{sec:CM}

\begin{figure}[htbp]
\centering
\includegraphics[width=\linewidth]{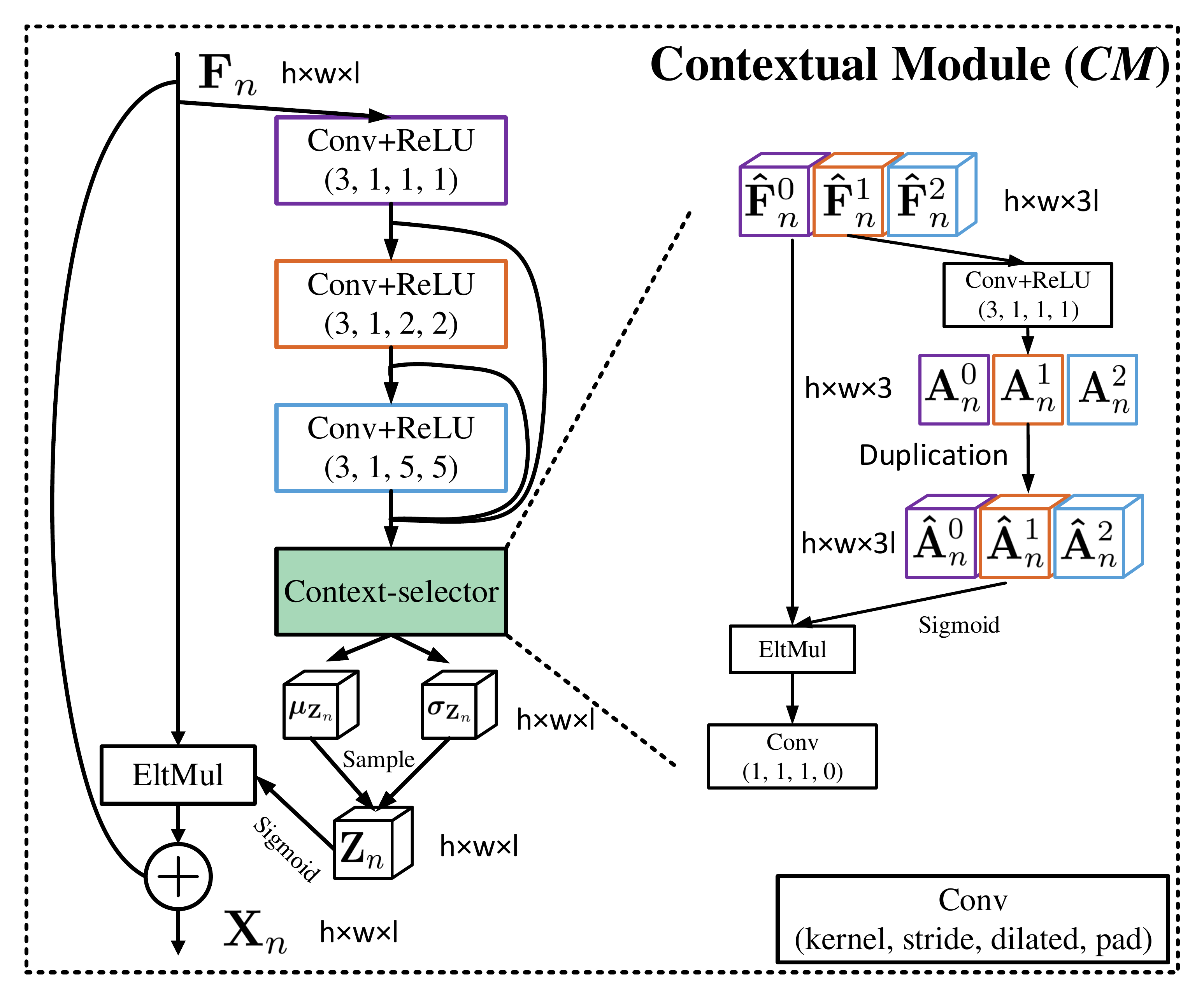}
\caption{\textbf{Contextual Module.} We aggregate the contextual information of different scales using our context selector. Then, the aggregated contextual information produces latent distribution for sampling contextual latent code. }
\label{contextual}
\end{figure}

\noindent\textbf{Multi-scale Context Maps:} We insert our Contextual Module ($CM$) after the backbone $E$ of Deeplabv2, as shown in Figure~\ref{ours}.
For the $n$-th image, we use $\textbf{F}_n\in \mathcal{R}^{h\times w\times l}$ to denote the output feature map of the $E$
Our $CM$ aims to gather the pixel-wise contextual information for each pixel on $\textbf{F}_n$.
Recall that the pixel-wise contextual information of a pixel means the aggregated information of its surrounding pixels. To achieve this goal, $CM$ takes $\textbf{F}_n$ as input to produce one or more context maps of the same size as $\textbf{F}_n$. Each pixel-wise vector on context maps contains the pixel-wise contextual information of its corresponding pixel on $\textbf{F}_n$.
In terms of the detailed design of $CM$, we consider two principles: 1) multi-scale contexts should be preserved for better feature generation; 2) the one-to-one correspondence between contexts and pixels should be maintained as discussed in Section~\ref{intro}, which means that no pooling layers should be used. Based on these principles, we employ several dilated conv layers~\cite{Yu2015Multi} because they support the exponential expansion of receptive fields without loss of spatial resolution.

As shown in Figure~\ref{contextual}, we use three serial dilated convs and refer to the output context maps of these layers as ${\hat{\textbf{F}}}^0_n,{\hat{\textbf{F}}}^1_n,{\hat{\textbf{F}}}^2_n\in \mathcal{R}^{h\times w\times l}$ respectively. 
Applying three successive context maps can capture contextual information of different scales because pixels on a deeper context map have larger receptive fields, which means information within larger neighborhoods can be collected for these pixels.


\noindent\textbf{Context Selector:} Next, we attempt to aggregate three context maps. Intuitively, the features of different pixels may be dominated by the contextual information of small receptive field (\emph{e.g.}, the posture or inner parts of its belonging object) or large receptive field (\emph{e.g.}, distant background objects). To better select the contextual information of suitable scale for each pixel, we propose a light-weight context selector to adaptively learn different scale weights for different pixels. Specifically, we employ a $3\times 3$ conv layer to transform the concatenated $ [{\hat{\textbf{F}}}^0_n,{\hat{\textbf{F}}}^1_n,{\hat{\textbf{F}}}^2_n]$ to a $3$-channel scale weight map $\textbf{{A}}_n=[\textbf{A}_n^0,\textbf{A}_n^1,\textbf{A}_n^2]\in \mathcal{R}^{h\times w\times 3}$, in which the $k$-th channel $\textbf{A}_n^k$ contains the weights of all pixels for the $k$-th scale. Then, we duplicate each channel $\textbf{A}_n^k$ to $l$ channels to get ${\hat{\textbf{A}}}_n^k \in \mathcal{R}^{h\times w\times l}$ and obtain the weighted concatenation of three context maps $[{\hat{\textbf{F}}}^0_n\odot{\hat{\textbf{A}}}_n^0,{\hat{\textbf{F}}}^1_n\odot{\hat{\textbf{A}}}_n^1,{\hat{\textbf{F}}}^2_n\odot{\hat{\textbf{A}}}_n^2]\in \mathcal{R}^{h\times w\times 3l}$ with $\odot$ being Hadamard product. In this way, we select contexts of different scales pixel-wisely.
Although our contextual module looks similar to channel attention ~\cite{SEnet,li2018harmonious} or full attention~\cite{wang2018mancs}, our motivation and technical details are intrinsically different from them. 

\noindent\textbf{Contextual Latent Code:} To obtain contextual latent code, we apply a $1\times 1$ conv layer to the weighted concatenation of context maps $[{\hat{\textbf{F}}}^0_n\odot{\hat{\textbf{A}}}_n^0,{\hat{\textbf{F}}}^1_n\odot{\hat{\textbf{A}}}_n^1,{\hat{\textbf{F}}}^2_n\odot{\hat{\textbf{A}}}_n^2]$ to output $\bm{\mu}_{\textbf{Z}_n}\in \mathcal{R}^{h\times w\times l}$ and $\bm{\sigma}_{\textbf{Z}_n}\in \mathcal{R}^{h\times w\times l}$, in which $\bm{\mu}_{\textbf{z}_{n,i}}$ and $\bm{\sigma}_{\textbf{z}_{n,i}}$ represent for each pixel-wise vector respectively. Then, the contextual latent code $\textbf{z}_{n,i}$ for the $i$-th pixel can be sampled from Gaussian distribution $\mathcal{N}(\bm{\mu}_{\textbf{z}_{n,i}},\bm{\sigma}_{\textbf{z}_{n,i}})$ by using $\textbf{z}_{n,i}=\bm{\mu}_{\textbf{z}_{n,i}}+\epsilon \bm{\sigma}_{\textbf{z}_{n,i}}$, with epsilon being a random scalar sampled from $\mathcal{N}(0,1)$. To enable stochastic sampling during inference, we employ a KL-divergence loss to enforce $\mathcal{N}(\bm{\mu}_{\textbf{z}_{n,i}},\bm{\sigma}_{\textbf{z}_{n,i}})$ to be close to unit Gaussian distribution $\mathcal{N}(\textbf{0,1})$: $$\mathcal{L}_{KL} = \mathcal{D}_{KL}[\mathcal{N}(\bm{\mu}_{\textbf{z}_{n,i}},\bm{\sigma}_{\textbf{z}_{n,i}})||\mathcal{N}(\textbf{0,1})].$$

We assume that the pixel-wise contextual latent code encodes the contextual information of this pixel. For instance, given a pixel in a cat near a tree, its contextual latent code may encode its near local region in the cat, its relative location in the cat, the posture of the cat, background objects like the tree, \emph{etc}.

Furthermore, we aggregate all $\textbf{z}_{n,i}$ for the $n$-th image into latent code map $\textbf{Z}_n \in \mathcal{R}^{h\times w\times l}$. Inspired by~\cite{hu2018gather}, we element-multiply $\textbf{Z}_n$ to $\textbf{F}_n$ after applying sigmoid activation (denoted as $\phi$) as residual attention, that is, our $CM$ outputs the new feature map $\textbf{X}_n = \textbf{F}_n+\textbf{F}_n\odot \phi(\textbf{Z}_n) \in \mathcal{R}^{h\times w\times l}$ as both the target of feature generation and the input of classifier $C$. In this way, $CM$ can be jointly trained with segmentation network as a residual attention module.
Note that $CM$ could slightly enhance the output feature map ($\textbf{X}_n$ \emph{v.s.} $\textbf{F}_n$ of segmentation network, see Section~\ref{sec:ablation}), but the main goal of $CM$ is to facilitate feature generation.

\subsection{Context-aware Feature Generator}\label{sec:context_feature_generator}

In this section, we first introduce the feature generation pipeline for seen categories, because training images only have pixel-wise annotations of seen objects. Given an input image $\textbf{I}_n$, the backbone $E$, together with the Contextual Module $CM$, delivers real visual feature map $\textbf{X}_n^s$ with pixel-wise feature $\textbf{x}_{n,i}^s$ and contextual latent code map $\textbf{Z}_n \in \mathcal{R}^{h\times w\times l}$ with pixel-wise latent code $\textbf{z}_{n,i}$ as mentioned in Section~\ref{sec:CM}. For the $i$-th pixel on $\textbf{X}_n^s$, , we have the category label $c_{n,i}^s$, which can also be represented by a one-hot vector $\textbf{y}_{n,i}^s$ from the segmentation label map $\textbf{Y}_n^s$. Note that $\textbf{Y}_n^s$ is a down-sampled label map with the same spatial resolution as $\textbf{X}_n^s$, \emph{i.e.}, $\textbf{Y}_n^s \in \mathcal{R}^{h\times w\times (|\mathcal{C}^s|+|\mathcal{C}^u|)}$. We can obtain the corresponding semantic word embedding map $\textbf{W}^s_n\in \mathcal{R}^{h\times w\times d}$ with pixel-wise category embedding $\textbf{w}^s_{n,i} = {\bar{\textbf{w}}}_{c_{n,i}^s}$. To generate fake pixel-wise feature, $\textbf{Z}_n$ is then concatenated with $\textbf{W}^s_n$ as the input of generator $G$, which can be written as ${\tilde{\textbf{x}}}_{n,i}^s=G(\textbf{z}_{n,i},\textbf{w}_{n,i}^s)$ for each pixel-wise generation process.
As discussed in Section~\ref{intro}, since category-specific $\textbf{w}^s_{n,i}$ and adequate contextual information $\textbf{z}_{n,i}$ is passed to $G$ to resolve the ambiguity of output, we expect $G$ to reconstruct the pixel-wise feature $\textbf{x}^s_{n,i}$. This goal is accomplished by a L2 reconstruction loss $\mathcal{L}_{REC}$:
\begin{eqnarray}
\mathcal{L}_{REC}=\sum_{n,i}||\textbf{x}^s_{n,i}-{\tilde{\textbf{x}}}^s_{n,i}||^2_2.
\end{eqnarray}
We also use a classification loss and an adversarial loss to regulate the generated features.
Since the down-sampled label map $\textbf{Y}_n^s$ has the same spatial resolution as the real feature map $\textbf{X}_n^s$, $\textbf{y}_{n,i}^s$ one-to-one corresponds to $\textbf{x}_{n,i}^s$ pixel-wisely. Following many segmentation papers~\cite{long2015fully,zhao2017pyramid,chen2018deeplab,Lin2016RefineNet}, we use the cross-entropy loss function as classification loss $\mathcal{L}_{CLS}$. It can be written as
\begin{eqnarray}\label{eqn:L_cls}
\mathcal{L}_{CLS} =-\sum_{n,i}\textbf{y}_{n,i}^s\log(C(\textbf{x}_{n,i}^s)),
\end{eqnarray}
where the segmentation score from $C$ is normalized by a softmax function.
Following~\cite{MaoLeast}, adversarial loss $\mathcal{L}_{ADV}$ can be written as
\begin{eqnarray}\label{eqn:L_adv}
\mathcal{L}_{ADV} =\sum_{n,i} (D(\textbf{x}_{n,i}^s))^2 + (1-D(G(\textbf{z}_{n,i}, \textbf{w}_{n,i}^s)))^2,
\end{eqnarray}
in which the discrimination score from $D$ is normalized within $[0,1]$ by a sigmoid function, with target $1$ (\emph{resp.}, $0$) indicating real (\emph{resp.}, fake) pixel-wise features.

Then, we introduce the pixel-wise feature generation pipeline for both seen and unseen categories. We can feed a latent code $\textbf{z}$ randomly sampled from $\mathcal{N}(0,1)$ and a semantic word embedding ${\bar{\textbf{w}}}_c$ into $G$ to generate a pixel-wise feature $G(\textbf{z}, {\bar{\textbf{w}}}_c)$ for arbitrary category $c\in \mathcal{C}^s\cup \mathcal{C}^u$. Intuitively, $G(\textbf{z}, {\bar{\textbf{w}}}_c)$ stands for the pixel-wise feature of category $c$ in the context encoded by $\textbf{z}$.

\subsection{Optimization}\label{sec:optimization}
As shown in Figure~\ref{ours}, our optimization procedure has two steps in different colors: training and finetuning.

1) \textbf{Training:} In this step, the segmentation network and the feature generation network are trained jointly based on image data and segmentation masks of only seen categories. All network modules $(E, CM, G, D, C)$ are updated. The objective function contains the loss terms introduced in Section~\ref{sec:context_feature_generator}:
$$ \min\limits_{G,E,C,CM}\max\limits_D \quad \mathcal{L}_{CLS}+\mathcal{L}_{ADV}+\lambda_1 \mathcal{L}_{REC}+\lambda_{2}\mathcal{L}_{KL}.$$
Note that during optimization, we first update the parameters in $ D $ by maximizing the objective function, aiming to improve the discrimination ability of $ D $. Then we try to minimize the objective function to update the other parameters of the network to both enhance the performances of segmentation and feature generation. 

2) \textbf{Finetuning:} In this step, we consider both seen and unseen categories, so that the segmentation network can generalize well to unseen categories. For ease of computation, we construct the $m$-th word embedding map ${\tilde{\textbf{W}}}^{s\cup u}_m\in \mathcal{R}^{h\times w\times d}$ by randomly stacking pixel-wise word embeddings ${\tilde{\textbf{w}}}^{s\cup u}_{m,i}$ of both seen and unseen categories. The corresponding label map is ${\tilde{\textbf{Y}}}^{s\cup u}_m$ with pixel-wise label vector ${\tilde{\textbf{y}}}^{s\cup u}_{m,i}$. We use approximately the same number of seen and unseen pixels in each ${\tilde{\textbf{W}}}^{s\cup u}_m$, which can generally achieves good performances as discussed in Table~\ref{training} of Section~\ref{Hyper-parameter_Analyses}. Then, we generate fake feature map ${\tilde{\textbf{X}}}^{s\cup u}_m$ with pixel-wise feature ${\tilde{\textbf{x}}}^{s\cup u}_{m,i}$, based on ${\tilde{\textbf{W}}}^{s\cup u}_m$ and contextual latent code map ${\tilde{\textbf{Z}}}_m$ with pixel-wise latent code ${\tilde{\textbf{z}}}_{m,i}$ sampled from $\mathcal{N}(\textbf{0,1})$. The above pixel-wise feature generation process can be formulated as ${\tilde{\textbf{x}}}^{s\cup u}_{m,i}=G({\tilde{\textbf{z}}}_{m,i},{\tilde{\textbf{w}}}^{s\cup u}_{m,i})$. We freeze $E$ and $CM$ because there are no real visual features for gradient backpropagation. Only $G$, $D$, and $C$ are updated. Thus, the objective function can be written as
$$ \min\limits_{G,C}\max\limits_D \quad \mathcal{\tilde{L}}_{CLS}+\mathcal{\tilde{L}}_{ADV},$$
in which $\mathcal{\tilde{L}}_{CLS}$ is obtained by replacing $\textbf{y}_{n,i}^s$ (\emph{resp.}, $\textbf{x}_{n,i}^s$) in (\ref{eqn:L_cls}) with ${\tilde{\textbf{y}}}_{m,i}^{s\cup u}$ (\emph{resp.}, ${\tilde{\textbf{x}}}_{m,i}^{s\cup u}$). For $\mathcal{\tilde{L}}_{ADV}$, we replace $\textbf{w}_{n,i}^s$ (\emph{resp.}, $\textbf{z}_{n,i}$) in (\ref{eqn:L_adv}) with ${\tilde{\textbf{w}}}_{m,i}^{s\cup u}$ (\emph{resp.}, ${\tilde{\textbf{z}}}_{m,i}$), and delete the first term $(D(\textbf{x}_{n,i}^s))^2$ within the summation formula since we have no real features in this step.
The optimizing process is the same as the training step by iteratively maximizing and minimizing the objective function.

\begin{table*}
\centering
\setlength{\tabcolsep}{1mm}{
\begin{tabular}{c|cccc|ccc|ccc}
\toprule[1.5pt]
\multicolumn{11}{c}{Pascal-Context}\\ \hline
\multirow{2}*{Method} & \multicolumn{4}{c}{Overall}   & \multicolumn{3}{c}{Seen}    & \multicolumn{3}{c}{Unseen}      \\ 
~     & \textbf{hIoU} & mIoU &  pixel acc.& mean acc. & mIoU  & pixel acc. &  mean acc. & mIoU  & pixel acc. &  mean acc. \\ \hline
SPNet& 0& 0.2938& 0.5793& 0.4486& 0.3357& \textbf{0.6389}& 0.5105& 0&0& 0\\ 
SPNet-c& 0.0718&0.3079& 0.5790& 0.4488& 0.3514&  0.6213& 0.4915&  0.0400& 0.1673&  0.1361\\ 
ZS3Net&0.1246  & 0.3010 &0.5710   & 0.4442 & 0.3304  &   0.6099  & 0.4843 & 0.0768 & 0.1922    &  0.1532   \\ 
\textbf{CaGNet}&\textbf{0.2061}  &\textbf{0.3347}  & \textbf{0.5924}  &\textbf{0.4900} & \textbf{0.3610} & 0.6189    & \textbf{0.5140} & \textbf{0.1442} & \textbf{0.3341}    &\textbf{0.3161}     \\ \hline
ZS3Net+ST& 0.1488 & 0.3102 & 0.5725  & 0.4532 & 0.3398 & 0.6107    &  0.4935&0.0953  &  0.2030  &   0.1721  \\
\textbf{CaGNet+ST} &\textbf{0.2252}  &\textbf{0.3352}   & \textbf{0.5961}  &\textbf{0.4962}  &\textbf{0.3644}   & \textbf{ 0.6120}    &\textbf{0.5065}  & \textbf{0.1630}  &  \textbf{0.4038}    &   \textbf{0.4214 } \\
\toprule[1.2pt]
\multicolumn{11}{c}{COCO-stuff}\\ \hline
SPNet& 0.0140 & 0.3164 & 0.5132  &  0.4593&0.3461  &   \textbf{0.6564}  & 0.5030 &0.0070  & 0.0171    &  0.0007   \\ 
SPNet-c&0.1398  & 0.3278 &0.5341   &0.4363  & 0.3518 &  0.6176   &0.4628  & 0.0873 &0.2450     & 0.1614    \\ 
ZS3Net& 0.1495 & 0.3328 & 0.5467  &0.4837  &0.3466  & 0.6434    & 0.5037 & 0.0953 & 0.2275    &   0.2701  \\ 
\textbf{CaGNet}&  \textbf{0.1819} &\textbf{0.3345}  & \textbf{0.5658}  &\textbf{0.4845}  & \textbf{0.3549} &  0.6562   & \textbf{0.5066} & \textbf{0.1223} & \textbf{0.2545}    &  \textbf{0.2701}   \\ \hline
ZS3Net+ST& 0.1620 &0.3367  & 0.5631  & \textbf{0.4862} &0.3489  & 0.6584    & 0.5042 & 0.1055  & 0.2488  &0.2718     \\
\textbf{CaGNet+ST} &\textbf{0.1946}  &\textbf{0.3372}   & \textbf{0.5676}  &0.4854  &\textbf{0.3555}   & \textbf{0.6587 }    &\textbf{0.5058}  & \textbf{0.1340}  &  \textbf{0.2670}    &   \textbf{0.2728}     \\ \toprule[1.2pt]
\multicolumn{11}{c}{Pascal-VOC}\\ \hline
SPNet& 0.0002 &0.5687 & 0.7685  & 0.7093 &0.7583  & \textbf{0.9482}    & \textbf{0.9458} & 0.0001 &   0.0007  &   0.0001  \\ 
SPNet-c&0.2610  &0.6315  & 0.7755  &0.7188  &0.7800  &0.8877     &0.8791  & 0.1563 &0.2955 &0.2387 \\ 
ZS3Net&0.2874  &0.6164  &0.7941   &0.7349  &0.7730  & 0.9296    &0.8772  &0.1765  &0.2147     &0.1580     \\ 
\textbf{CaGNet}&\textbf{0.3972}  &\textbf{0.6545}   & \textbf{0.8068}   &\textbf{0.7636}   & \textbf{0.7840}  & 0.8950     & 0.8868  &\textbf{0.2659}   & \textbf{0.4297}     &  \textbf{0.3940}    \\ \hline
ZS3Net+ST& 0.3328 & 0.6302 & 0.8095  &  0.7382&0.7802 &\textbf{0.9189}  &\textbf{0.8569}       & 0.2115 &  0.3407  & 0.2637    \\
\textbf{CaGNet+ST} &\textbf{0.4366}  &\textbf{0.6577}   & \textbf{0.8164}  &\textbf{0.7560}  &\textbf{0.7859}   & 0.8704    &0.8390  & \textbf{0.3031}  &  \textbf{0.5855}    &   \textbf{0.5071}     \\ \bottomrule[1.5pt]
\end{tabular}
\caption{Zero-shot segmentation performances on Pascal-Context, COCO-stuff and Pascal-VOC. ``ST'' stands for self-training. The best results with or w/o self-training are denoted in boldface, respectively.}
\label{coco}
}
\end{table*}

By using ResNet-101~\cite{resnet-101} pre-trained on ImageNet~\cite{ILSVRC15} as the initialization of backbone $E$, we first apply the training step on our network for enough iterations.
Next, we iteratively perform training and finetuning steps every 100 iterations to balance the network optimization based on real features and fake features. In the testing stage, a test image goes through segmentation backbone $ E $ and Contextual Module $ CM $ to obtain its real visual feature map, which is passed to classifier $ C $ to achieve segmentation results.

\section{Experiments}

\subsection{Datasets and Semantic Embeddings}
We evaluate our model on three benchmark datasets: Pascal-Context \cite{mottaghi2014role}, COCO-stuff~\cite{caesar2018coco}, and Pascal-VOC 2012~\cite{EveringhamThePascalVisual}. The Pascal-Context dataset contains 4998 training and 5105 validation images of 33 object/stuff categories. COCO-stuff has 164K images with dense pixel-wise annotations from 182 categories. Pascal-VOC 2012 contains 1464 training images with segmentation annotations of 20 object categories. For Pascal-VOC, following ZS3Net and SPNet, we adopt additional supervision from semantic boundary annotations~\cite{Hariharan2011Semantic} during training.

The experiment settings of two previous works, \emph{i.e.}, SPNet and ZS3Net, are different in many aspects (\emph{e.g.}, dataset, seen/unseen category split, backbone, semantic word embedding, evaluation metrics). SPNet reports results on large-scale COCO-stuff dataset\cite{caesar2018coco}, which makes their results more convincing. Besides, ZS3Net uses the word embedding of ``background'' as the semantic representation of all categories (\emph{e.g.}, sky and ground) belonging to ``background", which seems a little unreasonable, while SPNet ignores ``background" in both training and validation.
Thus, we choose to strictly follow the settings of SPNet. But we also report the results by strictly following the settings of ZS3Net in the supplementary. 

Following SPNet~\cite{xian2019semantic}, we concatenate two different types of word embeddings ($d=600$, $300$ for each), \emph{i.e.}, word2vec~\cite{mikolov2013distributed} trained on Google News and fast-Text~\cite{joulin2017bag} trained on Common Crawl. The word embeddings of categories that contain multiple words are obtained by averaging the embeddings of each individual word.

Our training/test sets are based on the standard train/test splits of three datasets, but we only use the pixel-wise annotations of seen categories and ignore the unseen pixels during training.
For seen/unseen category split, following SPNet, we treat ``frisbee, skateboard, cardboard, carrot, scissors, suitcase, giraffe, cow, road, wallconcrete, tree, grass, river, clouds, playingfield'' as $15$ unseen categories on COCO-stuff, and treat ``potted plant, sheep, sofa, train, tv monitor'' as $5$ unseen categories on Pascal-VOC. We additionally report results on Pascal-Context with $33$ categories, which is a popular segmentation dataset but not used in~\cite{xian2019semantic}. On Pascal-Context, we treat ``cow, motorbike, sofa, cat'' as $4$ unseen categories. 




\subsection{Implementation Details}\label{hypers}


Our generator $G$ is a multi-layer perceptron ($512$ hidden neurons, Leaky ReLU and dropout for each layer). Our classifier $C$ and discriminator $D$ consist of two $1\times 1$ conv layers and share the same weights in the first conv layer. During training, the learning rate is initialized as $2.5e^{-4}$ and divided by $10$ when the loss stops decreasing. The size of the training batch is $8$ on one Tesla V100. All input images are $368$ in size. We set $\lambda_{1}=10,\lambda_{2}=100$ via cross-validation by splitting a set of validation categories from seen categories. The analyses of $\lambda_{1},\lambda_{2}$ can be found in the supplementary.
We report results based on three evaluation metrics, \emph{i.e.}, pixel accuracy, mean accuracy and mean Intersection over Union (mIoU) for both seen and unseen categories. Moreover, we also calculate the harmonic IoU (hIoU)~\cite{xian2019semantic} of all categories.

\subsection{Comparison with State-of-the-art}\label{Implement}

We compare our method with two baselines: SPNet~\cite{xian2019semantic} and ZS3Net \cite{bucher2019zero}. For a fair comparison, we use the same backbone Deeplabv2 as in \cite{xian2019semantic} for all methods.
We also report the results of SPNet-c which deducts the prediction scores of seen categories by a calibration factor. Besides, we additionally employ the Self-Training (ST) strategy in~\cite{bucher2019zero} for both ZS3Net and our method. Specifically, we tag unlabeled pixels in training images using the trained segmentation model and add them to the training set to finetune the segmentation model iteratively.
We do not compare with ZS3Net (GC) in~\cite{bucher2019zero}, because their used graph contexts are unavailable in our setting and also difficult to acquire in real-world applications.

Among evaluation metrics, ``IoU'' quantizes the overlap between predicted and ground-truth objects, which is more reliable than ``accuracy'' considering the integrity of objects. For ``overall'' evaluation, ``hIoU'' is more valuable than ``mIoU'', because seen categories often have much higher mIoUs and dominate overall results.

Experimental results are summarized in Table~\ref{coco}. For unseen and overall evaluation, our CaGNet achieves significant improvement over SPNet\footnote{Our reproduced results of SPNet on Pascal-VOC dataset are obtained using their released model and code with careful tuning, but still lower than their reported results.} and ZS3Net on all three datasets, especially \emph{w.r.t.} ``mIoU'' and ``hIoU''. For seen evaluation, our method underperforms SPNet in some cases, because SPNet almost segments all pixels as seen categories while our method sacrifices some seen pixels for much better unseen performance.

\subsection{Ablation Study}\label{sec:ablation}

We evaluate our CaGNet on Pascal-VOC for ablation studies. We only report four reliable evaluation metrics: hIoU, mIoU, seen IoU (S-mIoU), and unseen IoU (U-mIoU), as claimed in Section~\ref{Implement}.

\begin{table}[tbp]
\centering
\resizebox{\columnwidth}!{
\begin{tabular}{cccc|cccc}
\toprule[1.5pt]
$E$ \& $C$   &$G$ &$D$   & $CM$  &\textbf{hIoU} &mIoU&S-mIoU&U-mIoU    \\ \hline \hline
\checkmark&&& &0 &0.5687   & 0.7583 & 0\\
\checkmark&&&\checkmark&0  & 0.5689  &0.7599  & 0\\
\checkmark&\checkmark& &&0.2911  &0.6332  & 0.7633 &0.1798\\
\checkmark&\checkmark&\checkmark &&0.3105  &0.6387  & 0.7751 &0.1941 \\
\checkmark&\checkmark& \checkmark &\checkmark& \textbf{0.3972} & \textbf{0.6545} & \textbf{0.7840}&\textbf{0.2659} \\ \bottomrule[1.5pt]
\end{tabular}}
\caption{Ablation studies of different network modules on Pascal-VOC. S-mIoU (\emph{resp.}, U-mIoU) is the mIoU of seen (\emph{resp.}, unseen) categories.} 
\label{novel}
\end{table}

\begin{table}[tbp]
\centering
\resizebox{\columnwidth}!{
\begin{tabular}{cccc|cccc}
\toprule[1.5pt]
Layer & Dilated   & MS & CS  &\textbf{hIoU} &mIoU&S-mIoU&U-mIoU    \\ \hline \hline
conv ($1\times 1$)&&&& 0.3211 & 0.6394  &0.7762  &0.2023 \\
conv&&&&0.3298  & 0.6408 & 0.7768&0.2093 \\
conv&\checkmark&&&0.3654 & 0.6502 & 0.7789 &0.2386\\
conv&\checkmark&\checkmark&&0.3825 &0.6526 &0.7810 &0.2532\\
conv (mask) & \checkmark&\checkmark &\checkmark &0.3961 & 0.6538 & 0.7821& 0.2652\\
conv & \checkmark&\checkmark &$\circ$ &0.3902 & 0.6529 & 0.7816 &0.2600 \\
conv&\checkmark& \checkmark &\checkmark & \textbf{0.3972} & \textbf{0.6545} & \textbf{0.7840}&\textbf{0.2659} \\ \bottomrule[1.5pt]
\end{tabular}}
\caption{Ablation studies of different variants of the contextual module on Pascal-VOC.  ``MS'' and ``CS'' stand for Multi-Scale and Context Selector respectively.}
\label{architecture}
\end{table}

\noindent \textbf{Validation of network modules:} We validate the effectiveness of each module ($E$, $C$, $G$, $D$, $CM$) in our method. The results are reported in Table~\ref{novel}, from which we can see that simply applying $CM$ to the segmentation network only brings marginal improvement. Feature generation with $G, D$ significantly raises the performance of unseen categories due to the reduced gap between seen and unseen categories. Finally, our proposed Contextual Module ($CM$) achieves evident improvements \emph{w.r.t.} all metrics.


\noindent \textbf{Variants of contextual module:} We explore different architectures before predicting $\bm{\mu}_{\textbf{Z}_n}$ and $\bm{\sigma}_{\textbf{Z}_n}$ in our Contextual Module ($CM$) from simple to complex in Table~\ref{architecture}, in which the last row is our proposed $CM$. The first row simply utilizes two $1\times 1$ conv layers without capturing contextual information, and the bad performance shows the benefit of using contextual information.

The second row utilizes five standard conv layers (number of model parameters equal to our $CM$) to capture contextual information. The third row replaces the first three conv layers in the second row with three dilated conv layers as in our $CM$ and achieves better results, which shows the benefit of using dilated conv. Built upon the third row, the fourth row further concatenates multi-scale contextual information $[{\hat{\textbf{F}}}^0_n,{\hat{\textbf{F}}}^1_n,{\hat{\textbf{F}}}^2_n]$ as in our $CM$ and applies a $1\times1$ conv layer, but does not use the context selector. The fourth row is better than the third row but worse than the last row, which proves the advantage of aggregating multi-scale contextual information and adaptively weighting different scales for different pixels.

\begin{table}[tbp]
\centering
\begin{tabular}{c|cccc}
\toprule[1.5pt]
loss &\textbf{hIoU} & mIoU& S-mIoU& U-mIoU    \\ \hline \hline
w/o $\mathcal{L}_{KL}$   & 0.3772   & 0.6513&0.7801& 0.2487    \\
w/o   $\mathcal{L}_{GAN}$  & 0.3154  &0.6392& 0.7753& 0.1979  \\
w/o   $\mathcal{L}_{REC}$  & 0.2176  &0.6473 & 0.7835& 0.1263  \\ \hline
CaGNet  & \textbf{0.3972} & \textbf{0.6545} & \textbf{0.7840}&\textbf{0.2659}   \\\bottomrule[1.5pt]
\end{tabular}
\caption{Ablation studies of loss terms on Pascal-VOC.}
\label{losses}
\end{table}

\begin{table}[tbp]
\centering
\begin{tabular}{c|cccc}
\toprule[1.5pt]
$r$ (seen:unseen) &\textbf{hIoU} & mIoU& S-mIoU& U-mIoU    \\ \hline \hline
 $|\mathcal{C}^s|:|\mathcal{C}^u|$   &0.2887  &0.6425&\textbf{0.7898}&  0.1763   \\
  $1:1$  &\textbf{0.3972} & \textbf{0.6545} & 0.7840&\textbf{0.2659}      \\
  $1:10$  &0.3896    &0.6375&0.7687& 0.2617    \\
    $0:1$  &  0.3766  &0.6024 &0.6620 & 0.2632\\ \bottomrule[1.5pt]
\end{tabular}
\caption{Performances of different \textbf{feature generating ratio} $r$ during the finetuning step on Pascal-VOC.}
\label{training}
\end{table}

We also study a special case of our $CM$ in the fifth row named ``conv (mask)". The only difference is that we set the central $1\times 1\times l$ weights of all $3\times 3\times l$ conv filters as constant zeros without any update in the first dilated conv layer.
In this way, when gathering the contextual information for each pixel, we roughly eliminate the impact of its own pixel-wise feature. The results in the fifth row are comparable with those in the last row, so it does not matter whether to eliminate the pixel-wise information for each pixel themselves.

\begin{figure}[tbp]
\centering
\includegraphics[width=\linewidth]{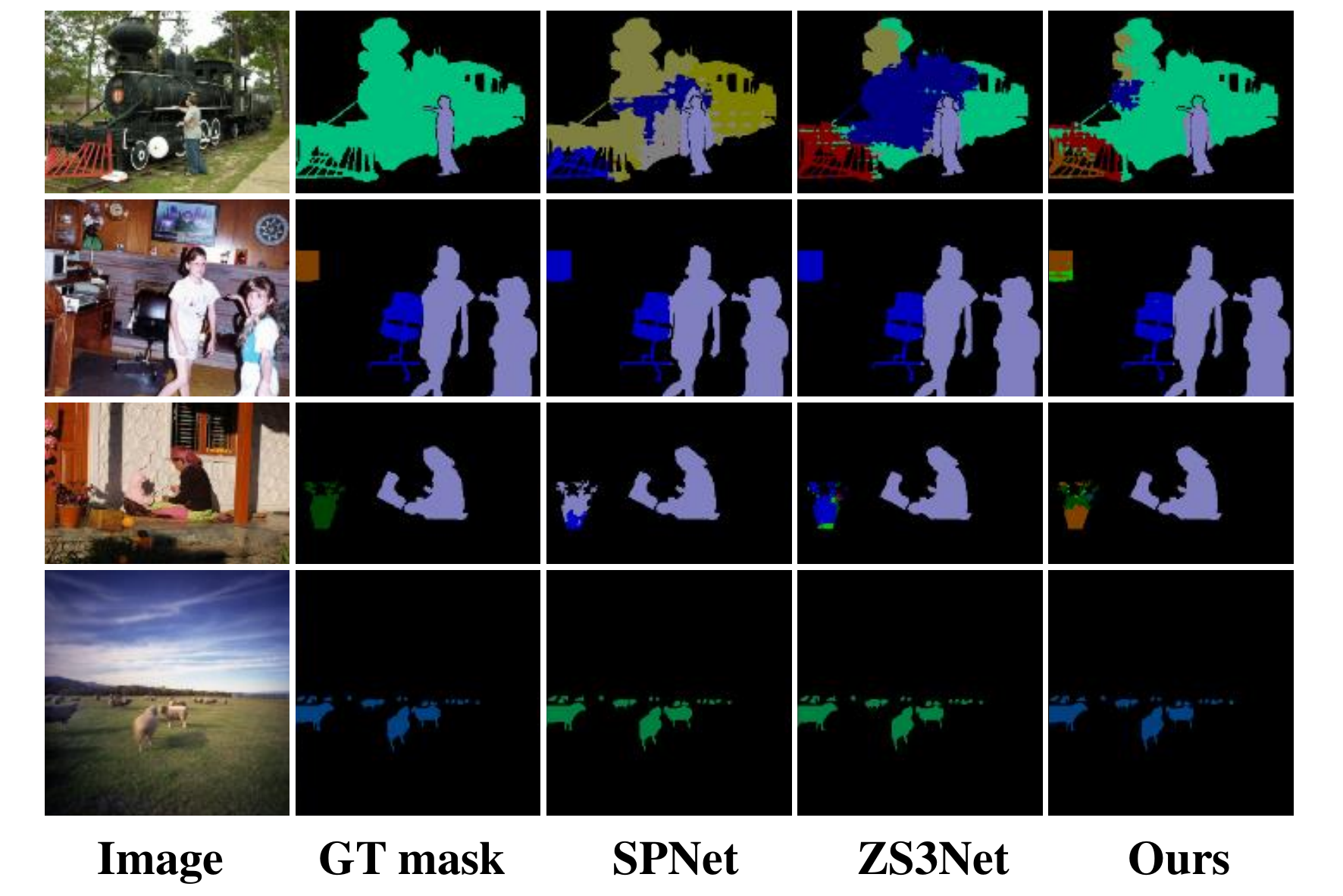}
\caption{Visualization of zero-shot segmentation results on Pascal-VOC. GT mask is ground-truth segmentation mask.}
\label{visseg}
\end{figure}

Another variant of our $CM$ is in the sixth row, in which $\circ$ means we modify the context selector by learning only one weight for each scale without considering inter-pixel difference. Specifically, we perform global average pooling on $[{\hat{\textbf{F}}}^0_n,{\hat{\textbf{F}}}^1_n,{\hat{\textbf{F}}}^2_n]$ followed by a FC to obtain a $1\times 1\times3$ scale weight vector, which is replicated to an $h\times w\times3l$ scale weight map. The results in the sixth row are also worse than the last row, showing the effectiveness of contextual information. Besides these results, we also evaluate two more special cases of $CM$, ``w/o residual'' and ``Parallel'' in the supplementary.

\noindent \textbf{Validation of loss terms:} We remove each loss term (\emph{i.e.}, $\mathcal{L}_{KL}$, $\mathcal{L}_{ADV}$, $\mathcal{L}_{REC}$) and report the results in Table~\ref{losses}. We observe that the performance becomes worse after removing any loss, which demonstrates that all loss terms contribute to better performance.



\subsection{Hyper-parameter Analysis}\label{Hyper-parameter_Analyses}


There exists a hyper-parameter during the finetuning step. We name it as the feature generating ratio (notated as $r$), which is the expected count ratio of seen pixels to unseen pixels while constructing each semantic word embedding map for feature generation. For example, if we randomly construct a word embedding map without any constraint, then $r=|\mathcal{C}^s|:|\mathcal{C}^u|$. However, in this case, seen features are much more than unseen features in pixel level (29:4 on Pascal-Context, 167:15 on COCO-stuff), leading to bad performances on unseen categories. After a few trials, we find that the reasonable feature generating ratio $r$ is $1:1$, as shown in Table~\ref{training}. The analyses of the other two hyper-parameters $\lambda_{1},\lambda_{2}$ can be found in the supplementary.

\begin{figure}[tbp]
\centering
\includegraphics[width=\linewidth]{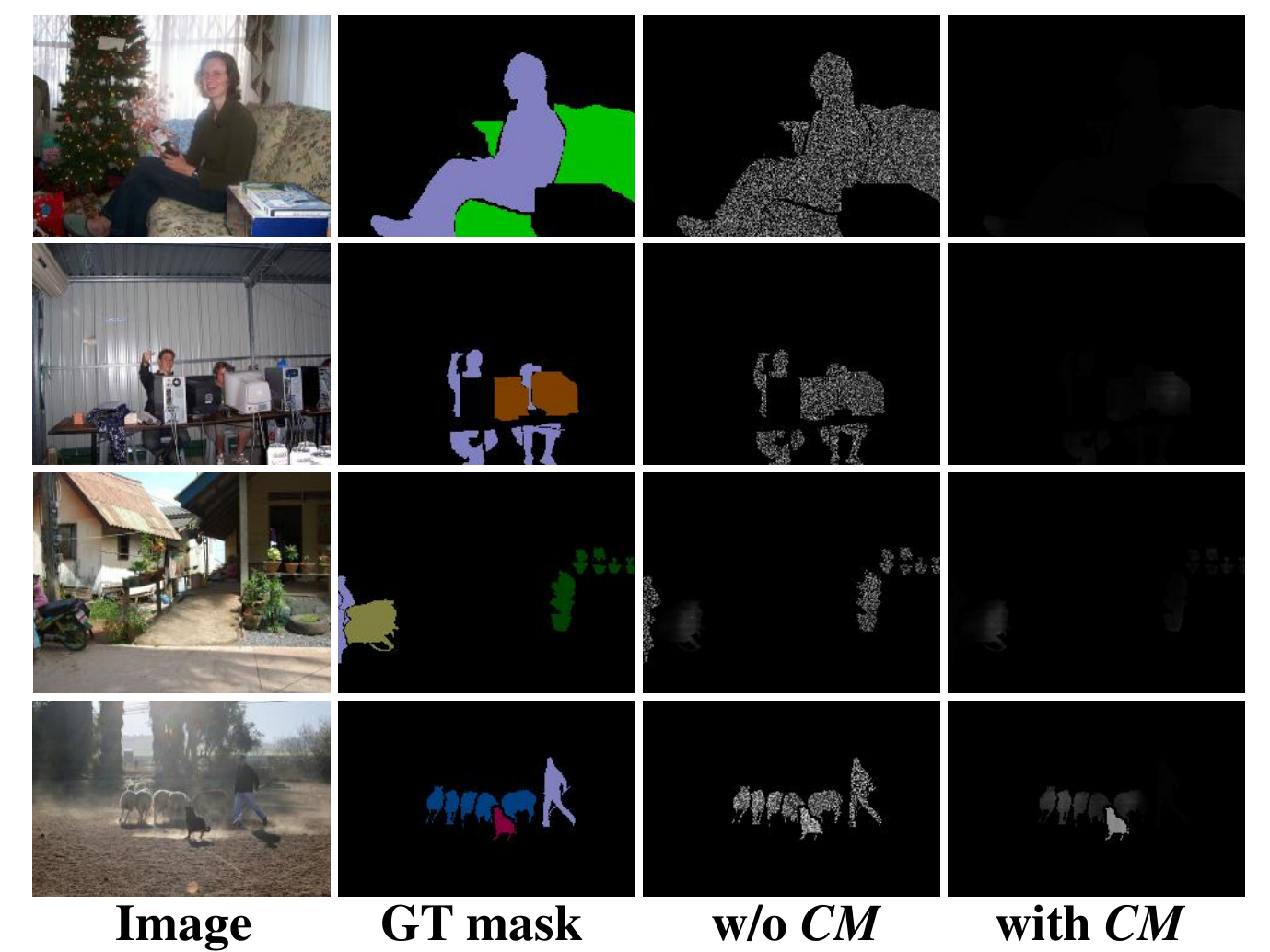}
\caption{Visualization of context-aware feature generation on Pascal-VOC test set. GT mask is ground-truth segmentation mask. In the third and fourth columns, we show the reconstruction loss maps calculated based on the generated feature maps and real feature maps (the darker, the better).}
\label{visgen}
\end{figure}

\subsection{Qualitative Analyses}

We also provide some visualizations on Pascal-VOC. More visualization results can be found in the Supplementary.

\noindent\textbf{Semantic segmentation:} We show the segmentation results of baselines and our method in Figure~\ref{visseg}, in which ``GT'' means ground-truth. Our method performs more favorably when segmenting unseen objects, \emph{e.g.}, the train (green), tv monitor (orange), potted plant (green), sheep (dark blue) in the sorted four rows.

\noindent \textbf{Feature generation:} To confirm the effectiveness of feature generation with Contextual Module ($CM$), we evaluate the generated features on test images. On the one hand, we feed ground-truth semantic word embeddings and latent code into the generator to obtain the generated feature map. On the other hand, we input the test image to the segmentation backbone to obtain the real feature map. Then, we show the reconstruction loss map calculated based on the generated and real feature maps in Figure~\ref{visgen}, in which smaller loss implies better generation quality. We compare our method ``with $CM$'' (latent code is contextual latent code produced by $CM$) with the special case ``w/o $CM$'' (latent code is random vector). We can observe that our $CM$ not only helps generate better features for seen categories (\emph{e.g.}, ``person''), but also for unseen categories (\emph{e.g.}, ``potted plant, sheep, sofa, tv monitor'').

\begin{figure}[tbp]
\centering
\includegraphics[width=0.9\linewidth]{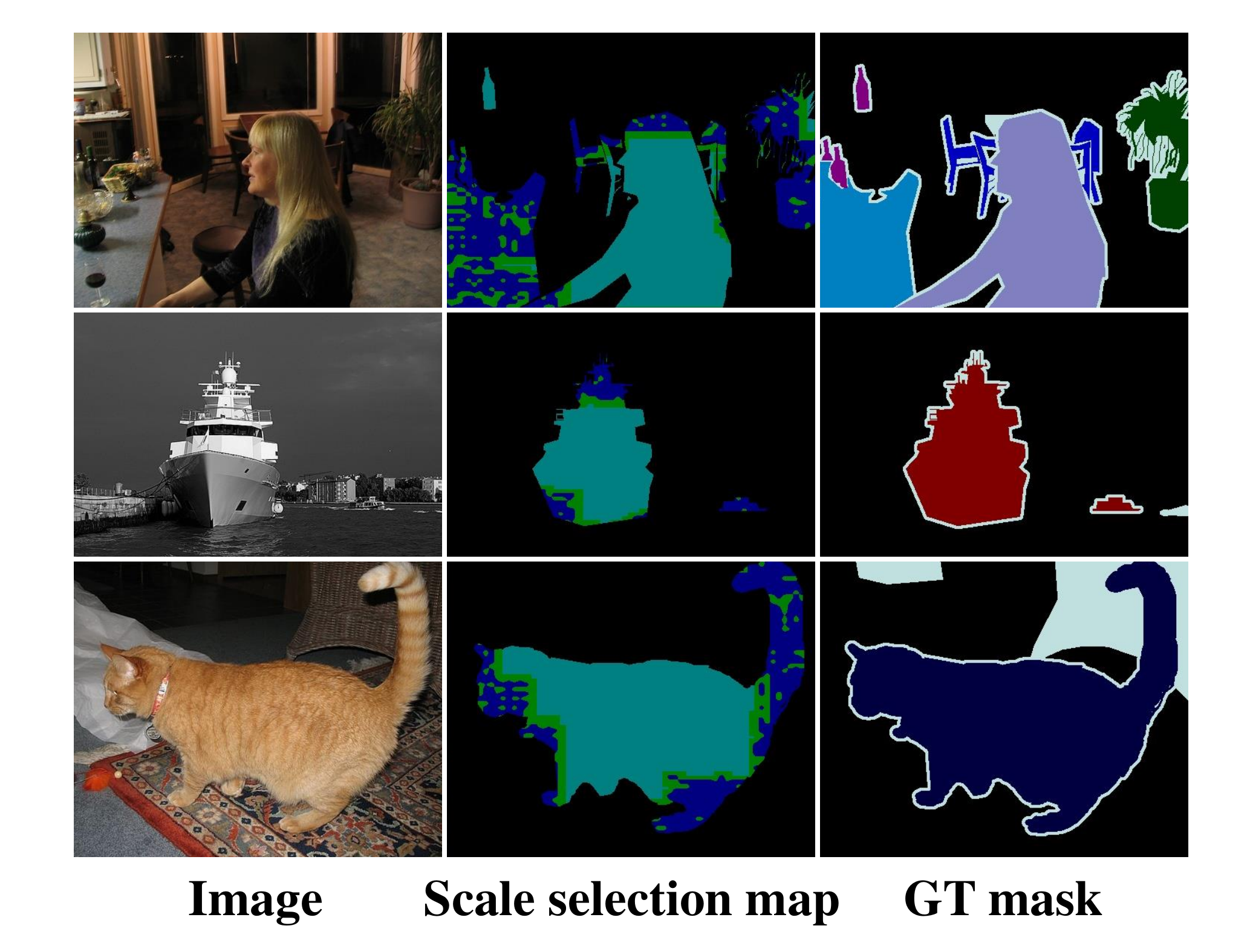}
\caption{Visualization of the effectiveness of context selector on Pascal-VOC. 
GT mask is ground-truth segmentation mask.
The scale selection map is obtained from the scale weight map, in which dark blue, green, light blue represents small scale, middle scale, and large scale respectively.}
\label{viscon}
\end{figure}
\noindent\textbf{Context selector:} The target of our context selector is to select context of the suitable scale for each pixel based on the scale weight map $[\textbf{A}_n^0,\textbf{A}_n^1,\textbf{A}_n^2]\in\mathcal{R}^{h\times w\times 3}$, in which each pixel-wise vector contains three scale weights for the pixel it corresponds to. In our implementation,
$\textbf{A}_n^0$ (\emph{resp.}, $\textbf{A}_n^1$, $\textbf{A}_n^2$) corresponds to small scale (\emph{resp.}, middle scale, large scale) with the size of receptive field being $3\times3$ (\emph{resp.}, $7\times7$, $17\times17$) \emph{w.r.t.} the input feature map $\textbf{F}_n$.
In Figure~\ref{viscon}, we list some examples with their corresponding scale selection maps and ground-truth segmentation masks.
Note that the scale selection map is obtained from the scale weight map by choosing the scale with the largest weight for each pixel. We use three colors to indicate the most suitable scale (largest weight) of each pixel.
In detail, dark blue, green, and light blue represent small scale, medium scale, and large scale respectively. 
From Figure~\ref{viscon}, we can observe that the pixels within discriminative local regions prefer the small scale while the other pixels prefer medium or large scale, which can be explained as follows.
For the pixels within discriminative local regions (\emph{e.g.}, animal faces, small objects on the table), small-scale contextual information is sufficient for reconstructing pixel-wise features, while other pixels may require contextual information of larger scale. Another observation is small (\emph{resp.}, large) objects prefer small (\emph{resp.}, large) scale (\emph{e.g.},
the small boat and the large boat in the second row).
These observations verify our motivation and the effectiveness of our proposed context selector.

\section{Conclusion}

In this work, we have unified the segmentation network and feature generation for zero-shot semantic segmentation, which utilizes contextual information to generate diverse and context-aware features. Qualitative and quantitative results on three benchmark datasets have shown the effectiveness of our method. 

\begin{acks}
  The work is supported by the National Key R\&D Program of China (2018AAA0100704) and is partially sponsored by the National Natural Science Foundation of China (Grant No.61902247) and Shanghai Sailing Program (19YF1424400).
\end{acks}

\bibliographystyle{ACM-Reference-Format}
\bibliography{ijcai20}


\begin{thebibliography}{53}


\ifx \showCODEN    \undefined \def \showCODEN     #1{\unskip}     \fi
\ifx \showDOI      \undefined \def \showDOI       #1{#1}\fi
\ifx \showISBNx    \undefined \def \showISBNx     #1{\unskip}     \fi
\ifx \showISBNxiii \undefined \def \showISBNxiii  #1{\unskip}     \fi
\ifx \showISSN     \undefined \def \showISSN      #1{\unskip}     \fi
\ifx \showLCCN     \undefined \def \showLCCN      #1{\unskip}     \fi
\ifx \shownote     \undefined \def \shownote      #1{#1}          \fi
\ifx \showarticletitle \undefined \def \showarticletitle #1{#1}   \fi
\ifx \showURL      \undefined \def \showURL       {\relax}        \fi
\providecommand\bibfield[2]{#2}
\providecommand\bibinfo[2]{#2}
\providecommand\natexlab[1]{#1}
\providecommand\showeprint[2][]{arXiv:#2}

\bibitem[\protect\citeauthoryear{Akata, Perronnin, Harchaoui, and Schmid}{Akata
  et~al\mbox{.}}{2015a}]%
        {akata2015label}
\bibfield{author}{\bibinfo{person}{Zeynep Akata}, \bibinfo{person}{Florent
  Perronnin}, \bibinfo{person}{Zaid Harchaoui}, {and} \bibinfo{person}{Cordelia
  Schmid}.} \bibinfo{year}{2015}\natexlab{a}.
\newblock \showarticletitle{Label-Embedding For Image Classification}.
\newblock \bibinfo{journal}{\emph{TPAMI}} \bibinfo{volume}{38},
  \bibinfo{number}{7} (\bibinfo{year}{2015}), \bibinfo{pages}{1425--1438}.
\newblock


\bibitem[\protect\citeauthoryear{Akata, Reed, Walter, Lee, and Schiele}{Akata
  et~al\mbox{.}}{2015b}]%
        {akata2015evaluation}
\bibfield{author}{\bibinfo{person}{Zeynep Akata}, \bibinfo{person}{Scott Reed},
  \bibinfo{person}{Daniel Walter}, \bibinfo{person}{Honglak Lee}, {and}
  \bibinfo{person}{Bernt Schiele}.} \bibinfo{year}{2015}\natexlab{b}.
\newblock \showarticletitle{Evaluation Of Output Embeddings For Fine-Grained
  Image Classification}. In \bibinfo{booktitle}{\emph{CVPR}}.
\newblock


\bibitem[\protect\citeauthoryear{Bucher, Vu, Cord, and P{\'e}rez}{Bucher
  et~al\mbox{.}}{2019}]%
        {bucher2019zero}
\bibfield{author}{\bibinfo{person}{Maxime Bucher}, \bibinfo{person}{Tuan-Hung
  Vu}, \bibinfo{person}{Mathieu Cord}, {and} \bibinfo{person}{Patrick
  P{\'e}rez}.} \bibinfo{year}{2019}\natexlab{}.
\newblock \showarticletitle{Zero-Shot Semantic Segmentation}. In
  \bibinfo{booktitle}{\emph{NeurIPS}}.
\newblock


\bibitem[\protect\citeauthoryear{Caesar, Uijlings, and Ferrari}{Caesar
  et~al\mbox{.}}{2018}]%
        {caesar2018coco}
\bibfield{author}{\bibinfo{person}{Holger Caesar}, \bibinfo{person}{Jasper
  Uijlings}, {and} \bibinfo{person}{Vittorio Ferrari}.}
  \bibinfo{year}{2018}\natexlab{}.
\newblock \showarticletitle{Coco-Stuff: Thing And Stuff Classes In Context}. In
  \bibinfo{booktitle}{\emph{CVPR}}.
\newblock


\bibitem[\protect\citeauthoryear{Chen, Papandreou, Kokkinos, Murphy, and
  Yuille}{Chen et~al\mbox{.}}{2018}]%
        {chen2018deeplab}
\bibfield{author}{\bibinfo{person}{Liang-Chieh Chen}, \bibinfo{person}{George
  Papandreou}, \bibinfo{person}{Iasonas Kokkinos}, \bibinfo{person}{Kevin
  Murphy}, {and} \bibinfo{person}{Alan~L Yuille}.}
  \bibinfo{year}{2018}\natexlab{}.
\newblock \showarticletitle{Deeplab: Semantic Image Segmentation With Deep
  Convolutional Nets, Atrous Convolution, And Fully Connected CRFs}.
\newblock \bibinfo{journal}{\emph{TPAMI}} \bibinfo{volume}{40},
  \bibinfo{number}{4} (\bibinfo{year}{2018}), \bibinfo{pages}{834--848}.
\newblock


\bibitem[\protect\citeauthoryear{Everingham, Eslami, Van~Gool, Williams, Winn,
  and Zisserman}{Everingham et~al\mbox{.}}{2015}]%
        {EveringhamThePascalVisual}
\bibfield{author}{\bibinfo{person}{Mark Everingham}, \bibinfo{person}{S.~M.~Ali
  Eslami}, \bibinfo{person}{Luc Van~Gool}, \bibinfo{person}{Christopher K.~I.
  Williams}, \bibinfo{person}{John Winn}, {and} \bibinfo{person}{Andrew
  Zisserman}.} \bibinfo{year}{2015}\natexlab{}.
\newblock \showarticletitle{ThePascalVisual Object Classes Challenge: A
  Retrospective}.
\newblock \bibinfo{journal}{\emph{IJCV}} \bibinfo{volume}{111},
  \bibinfo{number}{1} (\bibinfo{year}{2015}), \bibinfo{pages}{98--136}.
\newblock


\bibitem[\protect\citeauthoryear{Felix, Kumar, Reid, and Carneiro}{Felix
  et~al\mbox{.}}{2018}]%
        {FelixMulti}
\bibfield{author}{\bibinfo{person}{Rafael Felix}, \bibinfo{person}{B.~G.~Vijay
  Kumar}, \bibinfo{person}{Ian Reid}, {and} \bibinfo{person}{Gustavo
  Carneiro}.} \bibinfo{year}{2018}\natexlab{}.
\newblock \showarticletitle{Multi-Modal Cycle-Consistent Generalized Zero-Shot
  Learning}. In \bibinfo{booktitle}{\emph{ECCV}}.
\newblock


\bibitem[\protect\citeauthoryear{Frome, Corrado, Shlens, Bengio, Dean, Ranzato,
  and Mikolov}{Frome et~al\mbox{.}}{2013}]%
        {frome2013devise}
\bibfield{author}{\bibinfo{person}{Andrea Frome}, \bibinfo{person}{Greg~S
  Corrado}, \bibinfo{person}{Jon Shlens}, \bibinfo{person}{Samy Bengio},
  \bibinfo{person}{Jeff Dean}, \bibinfo{person}{Marc'Aurelio Ranzato}, {and}
  \bibinfo{person}{Tomas Mikolov}.} \bibinfo{year}{2013}\natexlab{}.
\newblock \showarticletitle{Devise: A Deep Visual-Semantic Embedding Model}. In
  \bibinfo{booktitle}{\emph{NeurIPS}}.
\newblock


\bibitem[\protect\citeauthoryear{Fu, Hospedales, Xiang, and Gong}{Fu
  et~al\mbox{.}}{2015}]%
        {fu2015transductive}
\bibfield{author}{\bibinfo{person}{Yanwei Fu}, \bibinfo{person}{Timothy~M
  Hospedales}, \bibinfo{person}{Tao Xiang}, {and} \bibinfo{person}{Shaogang
  Gong}.} \bibinfo{year}{2015}\natexlab{}.
\newblock \showarticletitle{Transductive Multi-View Zero-Shot Learning}.
\newblock \bibinfo{journal}{\emph{TPAMI}} \bibinfo{volume}{37},
  \bibinfo{number}{11} (\bibinfo{year}{2015}), \bibinfo{pages}{2332--2345}.
\newblock


\bibitem[\protect\citeauthoryear{Guo, Ding, Han, Shao, Lou, and Dai}{Guo
  et~al\mbox{.}}{2019}]%
        {DBLP:conf/ijcai/GuoDHSLD19}
\bibfield{author}{\bibinfo{person}{Yuchen Guo}, \bibinfo{person}{Guiguang
  Ding}, \bibinfo{person}{Jungong Han}, \bibinfo{person}{Hang Shao},
  \bibinfo{person}{Xin Lou}, {and} \bibinfo{person}{Qionghai Dai}.}
  \bibinfo{year}{2019}\natexlab{}.
\newblock \showarticletitle{Zero-Shot Learning With Many Classes By High-Rank
  Deep Embedding Networks}. In \bibinfo{booktitle}{\emph{IJCAI}}.
\newblock


\bibitem[\protect\citeauthoryear{Habibian, Mensink, and Snoek}{Habibian
  et~al\mbox{.}}{2014}]%
        {10.1145/2578726.2578746}
\bibfield{author}{\bibinfo{person}{Amirhossein Habibian},
  \bibinfo{person}{Thomas Mensink}, {and} \bibinfo{person}{Cees G.~M. Snoek}.}
  \bibinfo{year}{2014}\natexlab{}.
\newblock \showarticletitle{Composite Concept Discovery For Zero-Shot Video
  Event Detection}. In \bibinfo{booktitle}{\emph{ACMMM}}.
\newblock


\bibitem[\protect\citeauthoryear{Hariharan, Arbelaez, Bourdev, Maji, and
  Malik}{Hariharan et~al\mbox{.}}{2011}]%
        {Hariharan2011Semantic}
\bibfield{author}{\bibinfo{person}{Bharath Hariharan}, \bibinfo{person}{Pablo
  Arbelaez}, \bibinfo{person}{Lubomir~D. Bourdev}, \bibinfo{person}{Subhransu
  Maji}, {and} \bibinfo{person}{Jitendra Malik}.}
  \bibinfo{year}{2011}\natexlab{}.
\newblock \showarticletitle{Semantic Contours From Inverse Detectors}. In
  \bibinfo{booktitle}{\emph{ICCV}}.
\newblock


\bibitem[\protect\citeauthoryear{Hu, Shen, Albanie, Sun, and Vedaldi}{Hu
  et~al\mbox{.}}{2018}]%
        {hu2018gather}
\bibfield{author}{\bibinfo{person}{Jie Hu}, \bibinfo{person}{Li Shen},
  \bibinfo{person}{Samuel Albanie}, \bibinfo{person}{Gang Sun}, {and}
  \bibinfo{person}{Andrea Vedaldi}.} \bibinfo{year}{2018}\natexlab{}.
\newblock \showarticletitle{Gather-Excite: Exploiting Feature Context In
  Convolutional Neural Networks}. In \bibinfo{booktitle}{\emph{NeurIPS}}.
\newblock


\bibitem[\protect\citeauthoryear{Jie~Hu}{Jie~Hu}{2018}]%
        {SEnet}
\bibfield{author}{\bibinfo{person}{Gang~Sun Jie~Hu, Li~Shen}.}
  \bibinfo{year}{2018}\natexlab{}.
\newblock \showarticletitle{Squeeze-And-Excitation Networks}. In
  \bibinfo{booktitle}{\emph{CVPR}}.
\newblock


\bibitem[\protect\citeauthoryear{Joulin, Grave, Bojanowski, and Mikolov}{Joulin
  et~al\mbox{.}}{2017}]%
        {joulin2017bag}
\bibfield{author}{\bibinfo{person}{Armand Joulin}, \bibinfo{person}{Edouard
  Grave}, \bibinfo{person}{Piotr Bojanowski}, {and} \bibinfo{person}{Tomas
  Mikolov}.} \bibinfo{year}{2017}\natexlab{}.
\newblock \showarticletitle{Bag Of Tricks For Efficient Text Classification}.
  In \bibinfo{booktitle}{\emph{ECAL}}.
\newblock


\bibitem[\protect\citeauthoryear{Kaiming~He and Sun}{Kaiming~He and
  Sun}{2016}]%
        {resnet-101}
\bibfield{author}{\bibinfo{person}{Shaoqing~Ren Kaiming~He, Xiangyu~Zhang}
  {and} \bibinfo{person}{Jian Sun}.} \bibinfo{year}{2016}\natexlab{}.
\newblock \showarticletitle{Deep Residual Learning For Image Recognition}. In
  \bibinfo{booktitle}{\emph{CVPR}}.
\newblock


\bibitem[\protect\citeauthoryear{Kato, Yamasaki, and Aizawa}{Kato
  et~al\mbox{.}}{2019}]%
        {kato2019zero}
\bibfield{author}{\bibinfo{person}{Naoki Kato}, \bibinfo{person}{Toshihiko
  Yamasaki}, {and} \bibinfo{person}{Kiyoharu Aizawa}.}
  \bibinfo{year}{2019}\natexlab{}.
\newblock \showarticletitle{Zero-Shot Semantic Segmentation Via Variational
  Mapping}. In \bibinfo{booktitle}{\emph{ICCV Workshops}}.
\newblock


\bibitem[\protect\citeauthoryear{Khoreva, Benenson, Hosang, Hein, and
  Schiele}{Khoreva et~al\mbox{.}}{2017}]%
        {khoreva2017simple}
\bibfield{author}{\bibinfo{person}{Anna Khoreva}, \bibinfo{person}{Rodrigo
  Benenson}, \bibinfo{person}{Jan Hosang}, \bibinfo{person}{Matthias Hein},
  {and} \bibinfo{person}{Bernt Schiele}.} \bibinfo{year}{2017}\natexlab{}.
\newblock \showarticletitle{Simple Does It: Weakly Supervised Instance And
  Semantic Segmentation}. In \bibinfo{booktitle}{\emph{CVPR}}.
\newblock


\bibitem[\protect\citeauthoryear{Kipf and Welling}{Kipf and Welling}{2017}]%
        {kipf2016semi-supervised}
\bibfield{author}{\bibinfo{person}{Thomas Kipf} {and} \bibinfo{person}{Max
  Welling}.} \bibinfo{year}{2017}\natexlab{}.
\newblock \showarticletitle{Semi-Supervised Classification With Graph
  Convolutional Networks}.
\newblock \bibinfo{journal}{\emph{ICLR}}.
\newblock


\bibitem[\protect\citeauthoryear{Lampert, Nickisch, and Harmeling}{Lampert
  et~al\mbox{.}}{2009}]%
        {Lampert2009Learning}
\bibfield{author}{\bibinfo{person}{Christoph~H. Lampert},
  \bibinfo{person}{Hannes Nickisch}, {and} \bibinfo{person}{Stefan Harmeling}.}
  \bibinfo{year}{2009}\natexlab{}.
\newblock \showarticletitle{Learning To Detect Unseen Object Classes By
  Between-Class Attribute Transfer}. In \bibinfo{booktitle}{\emph{CVPR}}.
\newblock


\bibitem[\protect\citeauthoryear{Lampert, Nickisch, and Harmeling}{Lampert
  et~al\mbox{.}}{2013}]%
        {lampert2013attribute}
\bibfield{author}{\bibinfo{person}{Christoph~H Lampert},
  \bibinfo{person}{Hannes Nickisch}, {and} \bibinfo{person}{Stefan Harmeling}.}
  \bibinfo{year}{2013}\natexlab{}.
\newblock \showarticletitle{Attribute-Based Classification For Zero-Shot Visual
  Object Categorization}.
\newblock \bibinfo{journal}{\emph{TPAMI}} \bibinfo{volume}{36},
  \bibinfo{number}{3} (\bibinfo{year}{2013}), \bibinfo{pages}{453--465}.
\newblock


\bibitem[\protect\citeauthoryear{Li, Jin, Lu, Ding, Zhu, and Huang}{Li
  et~al\mbox{.}}{2019}]%
        {Li2019Leveraging}
\bibfield{author}{\bibinfo{person}{Jingjing Li}, \bibinfo{person}{Mengmeng
  Jin}, \bibinfo{person}{Ke Lu}, \bibinfo{person}{Zhengming Ding},
  \bibinfo{person}{Lei Zhu}, {and} \bibinfo{person}{Zi Huang}.}
  \bibinfo{year}{2019}\natexlab{}.
\newblock \showarticletitle{Leveraging The Invariant Side Of Generative
  Zero-Shot Learning}. In \bibinfo{booktitle}{\emph{CVPR}}.
\newblock


\bibitem[\protect\citeauthoryear{Li, Zhu, and Gong}{Li et~al\mbox{.}}{2018}]%
        {li2018harmonious}
\bibfield{author}{\bibinfo{person}{Wei Li}, \bibinfo{person}{Xiatian Zhu},
  {and} \bibinfo{person}{Shaogang Gong}.} \bibinfo{year}{2018}\natexlab{}.
\newblock \showarticletitle{Harmonious Attention Network For Person
  Re-Identification}. In \bibinfo{booktitle}{\emph{CVPR}}.
\newblock


\bibitem[\protect\citeauthoryear{Lin, Dai, Jia, He, and Sun}{Lin
  et~al\mbox{.}}{2016}]%
        {lin2016scribblesup}
\bibfield{author}{\bibinfo{person}{Di Lin}, \bibinfo{person}{Jifeng Dai},
  \bibinfo{person}{Jiaya Jia}, \bibinfo{person}{Kaiming He}, {and}
  \bibinfo{person}{Jian Sun}.} \bibinfo{year}{2016}\natexlab{}.
\newblock \showarticletitle{Scribblesup: Scribble-Supervised Convolutional
  Networks For Semantic Segmentation}. In \bibinfo{booktitle}{\emph{CVPR}}.
\newblock


\bibitem[\protect\citeauthoryear{Lin, Milan, Shen, and Reid}{Lin
  et~al\mbox{.}}{2017}]%
        {Lin2016RefineNet}
\bibfield{author}{\bibinfo{person}{Guosheng Lin}, \bibinfo{person}{Anton
  Milan}, \bibinfo{person}{Chunhua Shen}, {and} \bibinfo{person}{Ian Reid}.}
  \bibinfo{year}{2017}\natexlab{}.
\newblock \showarticletitle{RefineNet: Multi-Path Refinement Networks For
  High-Resolution Semantic Segmentation}. In \bibinfo{booktitle}{\emph{CVPR}}.
\newblock


\bibitem[\protect\citeauthoryear{Long, Shelhamer, and Darrell}{Long
  et~al\mbox{.}}{2015}]%
        {long2015fully}
\bibfield{author}{\bibinfo{person}{Jonathan Long}, \bibinfo{person}{Evan
  Shelhamer}, {and} \bibinfo{person}{Trevor Darrell}.}
  \bibinfo{year}{2015}\natexlab{}.
\newblock \showarticletitle{Fully Convolutional Networks For Semantic
  Segmentation}. In \bibinfo{booktitle}{\emph{CVPR}}.
\newblock


\bibitem[\protect\citeauthoryear{Long, Xu, Li, Shen, Song, and Shen}{Long
  et~al\mbox{.}}{2018}]%
        {10.1145/3240508.3240715}
\bibfield{author}{\bibinfo{person}{Teng Long}, \bibinfo{person}{Xing Xu},
  \bibinfo{person}{Youyou Li}, \bibinfo{person}{Fumin Shen},
  \bibinfo{person}{Jingkuan Song}, {and} \bibinfo{person}{Heng~Tao Shen}.}
  \bibinfo{year}{2018}\natexlab{}.
\newblock \showarticletitle{Pseudo Transfer With Marginalized Corrupted
  Attribute For Zero-Shot Learning}. In \bibinfo{booktitle}{\emph{ACMMM}}.
\newblock


\bibitem[\protect\citeauthoryear{Mandal, Narayan, Dwivedi, Gupta, Ahmed, Khan,
  and Shao}{Mandal et~al\mbox{.}}{2019}]%
        {Mandal2019Out}
\bibfield{author}{\bibinfo{person}{Devraj Mandal}, \bibinfo{person}{Sanath
  Narayan}, \bibinfo{person}{Saikumar Dwivedi}, \bibinfo{person}{Vikram Gupta},
  \bibinfo{person}{Shuaib Ahmed}, \bibinfo{person}{Fahad~Shahbaz Khan}, {and}
  \bibinfo{person}{Ling Shao}.} \bibinfo{year}{2019}\natexlab{}.
\newblock \showarticletitle{Out-Of-Distribution Detection For Generalized
  Zero-Shot Action Recognition}. In \bibinfo{booktitle}{\emph{CVPR}}.
\newblock


\bibitem[\protect\citeauthoryear{Mao, Li, Xie, Lau, Wang, and Smolley}{Mao
  et~al\mbox{.}}{2017}]%
        {MaoLeast}
\bibfield{author}{\bibinfo{person}{Xudong Mao}, \bibinfo{person}{Qing Li},
  \bibinfo{person}{Haoran Xie}, \bibinfo{person}{Raymond Y.~K. Lau},
  \bibinfo{person}{Zhen Wang}, {and} \bibinfo{person}{Stephen~Paul Smolley}.}
  \bibinfo{year}{2017}\natexlab{}.
\newblock \showarticletitle{Least Squares Generative Adversarial Networks}. In
  \bibinfo{booktitle}{\emph{ICCV}}.
\newblock


\bibitem[\protect\citeauthoryear{Mikolov, Sutskever, Chen, Corrado, and
  Dean}{Mikolov et~al\mbox{.}}{2013}]%
        {mikolov2013distributed}
\bibfield{author}{\bibinfo{person}{Tomas Mikolov}, \bibinfo{person}{Ilya
  Sutskever}, \bibinfo{person}{Kai Chen}, \bibinfo{person}{Greg~S Corrado},
  {and} \bibinfo{person}{Jeff Dean}.} \bibinfo{year}{2013}\natexlab{}.
\newblock \showarticletitle{Distributed Representations Of Words And Phrases
  And Their Compositionality}. In \bibinfo{booktitle}{\emph{NeurIPS}}.
\newblock


\bibitem[\protect\citeauthoryear{Mottaghi, Chen, Liu, Cho, Lee, Fidler,
  Urtasun, and Yuille}{Mottaghi et~al\mbox{.}}{2014}]%
        {mottaghi2014role}
\bibfield{author}{\bibinfo{person}{Roozbeh Mottaghi}, \bibinfo{person}{Xianjie
  Chen}, \bibinfo{person}{Xiaobai Liu}, \bibinfo{person}{Nam-Gyu Cho},
  \bibinfo{person}{Seong-Whan Lee}, \bibinfo{person}{Sanja Fidler},
  \bibinfo{person}{Raquel Urtasun}, {and} \bibinfo{person}{Alan Yuille}.}
  \bibinfo{year}{2014}\natexlab{}.
\newblock \showarticletitle{The Role Of Context For Object Detection And
  Semantic Segmentation In The Wild}. In \bibinfo{booktitle}{\emph{CVPR}}.
\newblock


\bibitem[\protect\citeauthoryear{Niu, Cai, Veeraraghavan, and Zhang}{Niu
  et~al\mbox{.}}{2019}]%
        {niu2019zero-shot}
\bibfield{author}{\bibinfo{person}{Li Niu}, \bibinfo{person}{Jianfei Cai},
  \bibinfo{person}{Ashok Veeraraghavan}, {and} \bibinfo{person}{Liqing Zhang}.}
  \bibinfo{year}{2019}\natexlab{}.
\newblock \showarticletitle{Zero-Shot Learning via Category-Specific
  Visual-Semantic Mapping and Label Refinement}.
\newblock \bibinfo{journal}{\emph{IEEE Transactions on Image Processing}}
  \bibinfo{volume}{28}, \bibinfo{number}{2} (\bibinfo{year}{2019}),
  \bibinfo{pages}{965--979}.
\newblock


\bibitem[\protect\citeauthoryear{Niu, Veeraraghavan, and Sabharwal}{Niu
  et~al\mbox{.}}{2018}]%
        {niu2018webly}
\bibfield{author}{\bibinfo{person}{Li Niu}, \bibinfo{person}{Ashok
  Veeraraghavan}, {and} \bibinfo{person}{Ashu Sabharwal}.}
  \bibinfo{year}{2018}\natexlab{}.
\newblock \showarticletitle{Webly Supervised Learning Meets Zero-shot Learning:
  A Hybrid Approach for Fine-Grained Classification}. In
  \bibinfo{booktitle}{\emph{CVPR}}.
\newblock


\bibitem[\protect\citeauthoryear{Oh, Benenson, Khoreva, Akata, Fritz, and
  Schiele}{Oh et~al\mbox{.}}{2017}]%
        {oh2017exploiting}
\bibfield{author}{\bibinfo{person}{Seong~Joon Oh}, \bibinfo{person}{Rodrigo
  Benenson}, \bibinfo{person}{Anna Khoreva}, \bibinfo{person}{Zeynep Akata},
  \bibinfo{person}{Mario Fritz}, {and} \bibinfo{person}{Bernt Schiele}.}
  \bibinfo{year}{2017}\natexlab{}.
\newblock \showarticletitle{Exploiting Saliency For Object Segmentation From
  Image Level Labels}. In \bibinfo{booktitle}{\emph{CVPR}}.
\newblock


\bibitem[\protect\citeauthoryear{Papandreou, Chen, Murphy, and
  Yuille}{Papandreou et~al\mbox{.}}{2015}]%
        {papandreou2015weakly}
\bibfield{author}{\bibinfo{person}{George Papandreou},
  \bibinfo{person}{Liang-Chieh Chen}, \bibinfo{person}{Kevin~P Murphy}, {and}
  \bibinfo{person}{Alan~L Yuille}.} \bibinfo{year}{2015}\natexlab{}.
\newblock \showarticletitle{Weakly-And Semi-Supervised Learning Of A Deep
  Convolutional Network For Semantic Image Segmentation}. In
  \bibinfo{booktitle}{\emph{ICCV}}.
\newblock


\bibitem[\protect\citeauthoryear{Pu, Huang, Guan, and Zou}{Pu
  et~al\mbox{.}}{2018}]%
        {GraphNet}
\bibfield{author}{\bibinfo{person}{Mengyang Pu}, \bibinfo{person}{Yaping
  Huang}, \bibinfo{person}{Qingji Guan}, {and} \bibinfo{person}{Qi Zou}.}
  \bibinfo{year}{2018}\natexlab{}.
\newblock \showarticletitle{GraphNet: Learning Image Pseudo Annotations For
  Weakly-Supervised Semantic Segmentation}. In
  \bibinfo{booktitle}{\emph{ACMMM}}.
\newblock


\bibitem[\protect\citeauthoryear{Romera-Paredes and Torr}{Romera-Paredes and
  Torr}{2015}]%
        {romera2015embarrassingly}
\bibfield{author}{\bibinfo{person}{Bernardino Romera-Paredes} {and}
  \bibinfo{person}{Philip Torr}.} \bibinfo{year}{2015}\natexlab{}.
\newblock \showarticletitle{An Embarrassingly Simple Approach To Zero-Shot
  Learning}. In \bibinfo{booktitle}{\emph{ICML}}.
\newblock


\bibitem[\protect\citeauthoryear{Ronneberger, Fischer, and Brox}{Ronneberger
  et~al\mbox{.}}{2015}]%
        {ronneberger2015u}
\bibfield{author}{\bibinfo{person}{Olaf Ronneberger}, \bibinfo{person}{Philipp
  Fischer}, {and} \bibinfo{person}{Thomas Brox}.}
  \bibinfo{year}{2015}\natexlab{}.
\newblock \showarticletitle{U-Net: Convolutional Networks For Biomedical Image
  Segmentation}. In \bibinfo{booktitle}{\emph{MICCAI}}.
\newblock


\bibitem[\protect\citeauthoryear{Russakovsky, Deng, Su, Krause, Satheesh, Ma,
  Huang, Karpathy, Khosla, Bernstein, Berg, and Fei-Fei}{Russakovsky
  et~al\mbox{.}}{2015}]%
        {ILSVRC15}
\bibfield{author}{\bibinfo{person}{Olga Russakovsky}, \bibinfo{person}{Jia
  Deng}, \bibinfo{person}{Hao Su}, \bibinfo{person}{Jonathan Krause},
  \bibinfo{person}{Sanjeev Satheesh}, \bibinfo{person}{Sean Ma},
  \bibinfo{person}{Zhiheng Huang}, \bibinfo{person}{Andrej Karpathy},
  \bibinfo{person}{Aditya Khosla}, \bibinfo{person}{Michael Bernstein},
  \bibinfo{person}{Alexander~C. Berg}, {and} \bibinfo{person}{Li Fei-Fei}.}
  \bibinfo{year}{2015}\natexlab{}.
\newblock \showarticletitle{ImageNet Large Scale Visual Recognition Challenge}.
\newblock \bibinfo{journal}{\emph{IJCV}} \bibinfo{volume}{115},
  \bibinfo{number}{3} (\bibinfo{year}{2015}), \bibinfo{pages}{211--252}.
\newblock


\bibitem[\protect\citeauthoryear{Sariyildiz and Cinbis}{Sariyildiz and
  Cinbis}{2019}]%
        {Mert2019Gradient}
\bibfield{author}{\bibinfo{person}{Mert~Bulent Sariyildiz} {and}
  \bibinfo{person}{Ramazan~Gokberk Cinbis}.} \bibinfo{year}{2019}\natexlab{}.
\newblock \showarticletitle{Gradient Matching Generative Networks For Zero-Shot
  Learning}. In \bibinfo{booktitle}{\emph{CVPR}}.
\newblock


\bibitem[\protect\citeauthoryear{Wang, Zhang, Huang, Liu, and Wang}{Wang
  et~al\mbox{.}}{2018}]%
        {wang2018mancs}
\bibfield{author}{\bibinfo{person}{Cheng Wang}, \bibinfo{person}{Qian Zhang},
  \bibinfo{person}{Chang Huang}, \bibinfo{person}{Wenyu Liu}, {and}
  \bibinfo{person}{Xinggang Wang}.} \bibinfo{year}{2018}\natexlab{}.
\newblock \showarticletitle{Mancs: A Multi-Task Attentional Network With
  Curriculum Sampling For Person Re-Identification}. In
  \bibinfo{booktitle}{\emph{ECCV}}.
\newblock


\bibitem[\protect\citeauthoryear{Xian, Sangkloy, Agrawal, Raj, Lu, Fang, Yu,
  and Hays}{Xian et~al\mbox{.}}{2018b}]%
        {xian2018texturegan}
\bibfield{author}{\bibinfo{person}{Wenqi Xian}, \bibinfo{person}{Patsorn
  Sangkloy}, \bibinfo{person}{Varun Agrawal}, \bibinfo{person}{Amit Raj},
  \bibinfo{person}{Jingwan Lu}, \bibinfo{person}{Chen Fang},
  \bibinfo{person}{Fisher Yu}, {and} \bibinfo{person}{James Hays}.}
  \bibinfo{year}{2018}\natexlab{b}.
\newblock \showarticletitle{Texturegan: Controlling Deep Image Synthesis With
  Texture Patches}. In \bibinfo{booktitle}{\emph{CVPR}}.
\newblock


\bibitem[\protect\citeauthoryear{Xian, Choudhury, He, Schiele, and Akata}{Xian
  et~al\mbox{.}}{2019a}]%
        {xian2019semantic}
\bibfield{author}{\bibinfo{person}{Yongqin Xian}, \bibinfo{person}{Subhabrata
  Choudhury}, \bibinfo{person}{Yang He}, \bibinfo{person}{Bernt Schiele}, {and}
  \bibinfo{person}{Zeynep Akata}.} \bibinfo{year}{2019}\natexlab{a}.
\newblock \showarticletitle{Semantic Projection Network For Zero-and Few-Label
  Semantic Segmentation}. In \bibinfo{booktitle}{\emph{CVPR}}.
\newblock


\bibitem[\protect\citeauthoryear{Xian, Lorenz, Schiele, and Akata}{Xian
  et~al\mbox{.}}{2018a}]%
        {xian2018feature}
\bibfield{author}{\bibinfo{person}{Yongqin Xian}, \bibinfo{person}{Tobias
  Lorenz}, \bibinfo{person}{Bernt Schiele}, {and} \bibinfo{person}{Zeynep
  Akata}.} \bibinfo{year}{2018}\natexlab{a}.
\newblock \showarticletitle{Feature Generating Networks For Zero-Shot
  Learning}. In \bibinfo{booktitle}{\emph{CVPR}}.
\newblock


\bibitem[\protect\citeauthoryear{Xian, Sharma, Schiele, and Akata}{Xian
  et~al\mbox{.}}{2019b}]%
        {xian2019f}
\bibfield{author}{\bibinfo{person}{Yongqin Xian}, \bibinfo{person}{Saurabh
  Sharma}, \bibinfo{person}{Bernt Schiele}, {and} \bibinfo{person}{Zeynep
  Akata}.} \bibinfo{year}{2019}\natexlab{b}.
\newblock \showarticletitle{f-VAEGAN-D2: A Feature Generating Framework For
  Any-Shot Learning}. In \bibinfo{booktitle}{\emph{CVPR}}.
\newblock


\bibitem[\protect\citeauthoryear{Yang, Luo, Chen, Shen, Shao, and Shen}{Yang
  et~al\mbox{.}}{2016}]%
        {10.1145/2964284.2964319}
\bibfield{author}{\bibinfo{person}{Yang Yang}, \bibinfo{person}{Yadan Luo},
  \bibinfo{person}{Weilun Chen}, \bibinfo{person}{Fumin Shen},
  \bibinfo{person}{Jie Shao}, {and} \bibinfo{person}{Heng~Tao Shen}.}
  \bibinfo{year}{2016}\natexlab{}.
\newblock \showarticletitle{Zero-Shot Hashing Via Transferring Supervised
  Knowledge}. In \bibinfo{booktitle}{\emph{ACMMM}}.
\newblock


\bibitem[\protect\citeauthoryear{Yao, Han, Gong, and Lei}{Yao
  et~al\mbox{.}}{2015}]%
        {Yao2015Semantic}
\bibfield{author}{\bibinfo{person}{Xiwen Yao}, \bibinfo{person}{Junwei Han},
  \bibinfo{person}{Cheng Gong}, {and} \bibinfo{person}{Guo Lei}.}
  \bibinfo{year}{2015}\natexlab{}.
\newblock \showarticletitle{Semantic Segmentation Based On Stacked
  Discriminative Autoencoders And Context-Constrained Weakly Supervised
  Learning}. In \bibinfo{booktitle}{\emph{ACMMM}}.
\newblock


\bibitem[\protect\citeauthoryear{Yu and Koltun}{Yu and Koltun}{2016}]%
        {Yu2015Multi}
\bibfield{author}{\bibinfo{person}{Fisher Yu} {and} \bibinfo{person}{Vladlen
  Koltun}.} \bibinfo{year}{2016}\natexlab{}.
\newblock \showarticletitle{Multi-Scale Context Aggregation By Dilated
  Convolutions}. In \bibinfo{booktitle}{\emph{ICLR}}.
\newblock


\bibitem[\protect\citeauthoryear{Zhao, Puig, Zhou, Fidler, and Torralba}{Zhao
  et~al\mbox{.}}{2017a}]%
        {zhao2017open}
\bibfield{author}{\bibinfo{person}{Hang Zhao}, \bibinfo{person}{Xavier Puig},
  \bibinfo{person}{Bolei Zhou}, \bibinfo{person}{Sanja Fidler}, {and}
  \bibinfo{person}{Antonio Torralba}.} \bibinfo{year}{2017}\natexlab{a}.
\newblock \showarticletitle{Open Vocabulary Scene Parsing}. In
  \bibinfo{booktitle}{\emph{ICCV}}.
\newblock


\bibitem[\protect\citeauthoryear{Zhao, Shi, Qi, Wang, and Jia}{Zhao
  et~al\mbox{.}}{2017b}]%
        {zhao2017pyramid}
\bibfield{author}{\bibinfo{person}{Hengshuang Zhao}, \bibinfo{person}{Jianping
  Shi}, \bibinfo{person}{Xiaojuan Qi}, \bibinfo{person}{Xiaogang Wang}, {and}
  \bibinfo{person}{Jiaya Jia}.} \bibinfo{year}{2017}\natexlab{b}.
\newblock \showarticletitle{Pyramid Scene Parsing Network}. In
  \bibinfo{booktitle}{\emph{CVPR}}.
\newblock


\bibitem[\protect\citeauthoryear{Zhu, Park, Isola, and Efros}{Zhu
  et~al\mbox{.}}{2017a}]%
        {zhu2017unpaired}
\bibfield{author}{\bibinfo{person}{Jun-Yan Zhu}, \bibinfo{person}{Taesung
  Park}, \bibinfo{person}{Phillip Isola}, {and} \bibinfo{person}{Alexei~A
  Efros}.} \bibinfo{year}{2017}\natexlab{a}.
\newblock \showarticletitle{Unpaired Image-To-Image Translation Using
  Cycle-Consistent Adversarial Networks}. In \bibinfo{booktitle}{\emph{ICCV}}.
\newblock


\bibitem[\protect\citeauthoryear{Zhu, Zhang, Pathak, Darrell, Efros, Wang, and
  Shechtman}{Zhu et~al\mbox{.}}{2017b}]%
        {NeurIPS2017_6650}
\bibfield{author}{\bibinfo{person}{Jun-Yan Zhu}, \bibinfo{person}{Richard
  Zhang}, \bibinfo{person}{Deepak Pathak}, \bibinfo{person}{Trevor Darrell},
  \bibinfo{person}{Alexei~A Efros}, \bibinfo{person}{Oliver Wang}, {and}
  \bibinfo{person}{Eli Shechtman}.} \bibinfo{year}{2017}\natexlab{b}.
\newblock \showarticletitle{Toward Multimodal Image-To-Image Translation}. In
  \bibinfo{booktitle}{\emph{NeurIPS}}.
\newblock


\bibitem[\protect\citeauthoryear{Ziheng~Zhang}{Ziheng~Zhang}{2019}]%
        {GaoLearning}
\bibfield{author}{\bibinfo{person}{Ling Xie Jingyi Yu Shenghua~Gao
  Ziheng~Zhang, Anpei~Chen}.} \bibinfo{year}{2019}\natexlab{}.
\newblock \showarticletitle{Learning Semantics-Aware Distance Map With
  Semantics Layering Network For Amodal Instance Segmentation}. In
  \bibinfo{booktitle}{\emph{ACMMM}}.
\newblock


\end{thebibliography}



\begin{thebibliography}{4}


\ifx \showCODEN    \undefined \def \showCODEN     #1{\unskip}     \fi
\ifx \showDOI      \undefined \def \showDOI       #1{#1}\fi
\ifx \showISBNx    \undefined \def \showISBNx     #1{\unskip}     \fi
\ifx \showISBNxiii \undefined \def \showISBNxiii  #1{\unskip}     \fi
\ifx \showISSN     \undefined \def \showISSN      #1{\unskip}     \fi
\ifx \showLCCN     \undefined \def \showLCCN      #1{\unskip}     \fi
\ifx \shownote     \undefined \def \shownote      #1{#1}          \fi
\ifx \showarticletitle \undefined \def \showarticletitle #1{#1}   \fi
\ifx \showURL      \undefined \def \showURL       {\relax}        \fi
\providecommand\bibfield[2]{#2}
\providecommand\bibinfo[2]{#2}
\providecommand\natexlab[1]{#1}
\providecommand\showeprint[2][]{arXiv:#2}

\bibitem[\protect\citeauthoryear{Bucher, Vu, Cord, and P{\'e}rez}{Bucher
  et~al\mbox{.}}{2019}]%
        {bucher2019zero}
\bibfield{author}{\bibinfo{person}{Maxime Bucher}, \bibinfo{person}{Tuan-Hung
  Vu}, \bibinfo{person}{Mathieu Cord}, {and} \bibinfo{person}{Patrick
  P{\'e}rez}.} \bibinfo{year}{2019}\natexlab{}.
\newblock \showarticletitle{Zero-Shot Semantic Segmentation}. In
  \bibinfo{booktitle}{\emph{NeurIPS}}.
\newblock


\bibitem[\protect\citeauthoryear{Liang-Chieh~Chen}{Liang-Chieh~Chen}{2018}]%
        {deeplabv3}
\bibfield{author}{\bibinfo{person}{George Papandreou Florian Schroff
  Hartwig~Adam Liang-Chieh~Chen, Yukun~Zhu}.} \bibinfo{year}{2018}\natexlab{}.
\newblock \showarticletitle{Encoder-Decoder With Atrous Separable Convolution
  For Semantic Image Segmentation}. In \bibinfo{booktitle}{\emph{ECCV}}.
\newblock


\bibitem[\protect\citeauthoryear{Mikolov, Sutskever, Chen, Corrado, and
  Dean}{Mikolov et~al\mbox{.}}{2013}]%
        {mikolov2013distributed}
\bibfield{author}{\bibinfo{person}{Tomas Mikolov}, \bibinfo{person}{Ilya
  Sutskever}, \bibinfo{person}{Kai Chen}, \bibinfo{person}{Greg~S Corrado},
  {and} \bibinfo{person}{Jeff Dean}.} \bibinfo{year}{2013}\natexlab{}.
\newblock \showarticletitle{Distributed Representations Of Words And Phrases
  And Their Compositionality}. In \bibinfo{booktitle}{\emph{NeurIPS}}.
\newblock


\bibitem[\protect\citeauthoryear{Xian, Choudhury, He, Schiele, and Akata}{Xian
  et~al\mbox{.}}{2019}]%
        {xian2019semantic}
\bibfield{author}{\bibinfo{person}{Yongqin Xian}, \bibinfo{person}{Subhabrata
  Choudhury}, \bibinfo{person}{Yang He}, \bibinfo{person}{Bernt Schiele}, {and}
  \bibinfo{person}{Zeynep Akata}.} \bibinfo{year}{2019}\natexlab{}.
\newblock \showarticletitle{Semantic Projection Network For Zero-and Few-Label
  Semantic Segmentation}. In \bibinfo{booktitle}{\emph{CVPR}}.
\newblock


\end{thebibliography}

\end{document}


\fancyhead{}
\title{Supplementary for Context-aware Feature Generation For Zero-shot Semantic Segmentation}
\author{Zhangxuan Gu}
\orcid{0000-0002-2102-2693}
\affiliation{\institution{\mbox{MoE Key Lab of Artificial Intelligence,} Shanghai Jiao Tong University}}
\email{zhangxgu@126.com}
\author{Siyuan Zhou}
\affiliation{\institution{\mbox{MoE Key Lab of Artificial Intelligence,} Shanghai Jiao Tong University}}
\email{ssluvble@sjtu.edu.cn}
\author{Li Niu*}
\affiliation{\institution{\mbox{MoE Key Lab of Artificial Intelligence,} Shanghai Jiao Tong University}}
\email{ustcnewly@sjtu.edu.cn}
\author{Zihan Zhao}
\affiliation{\institution{\mbox{MoE Key Lab of Artificial Intelligence,} Shanghai Jiao Tong University}}
\email{john745111625@gmail.com}
\author{Liqing Zhang*}
\affiliation{\institution{\mbox{MoE Key Lab of Artificial Intelligence,} Shanghai Jiao Tong University}}
\email{zhang-lq@cs.sjtu.edu.cn}
\thanks{*Corresponding authors.}

\renewcommand{\shortauthors}{Gu, et al.}

\ccsdesc[500]{Computing methodologies~Image segmentation}

\maketitle

\begin{table*}
\centering
\setlength{\tabcolsep}{1mm}{
\begin{tabular}{c|cccc|ccc|ccc}
\toprule[1.5pt]
\multicolumn{11}{c}{Pascal-Context}\\ \hline
\multirow{2}*{Method} & \multicolumn{4}{c}{Overall}   & \multicolumn{3}{c}{Seen}    & \multicolumn{3}{c}{Unseen}      \\ 
~     & \textbf{hIoU} & mIoU &  pixel acc.& mean acc. & mIoU  & pixel acc. &  mean acc. & mIoU  & pixel acc. &  mean acc. \\ \hline
ZS3Net&16.3  & 19.5 & 54.6  &27.1  &  20.7 & 53.9    & 23.8 &13.5 &  59.6   &   43.8  \\ 
\textbf{CaGNet}&\textbf{21.2}  &\textbf{23.2}  & \textbf{56.6}  &\textbf{36.8} & \textbf{24.8} &  \textbf{55.2}  & \textbf{35.7} & \textbf{18.5} & \textbf{66.8}    &\textbf{49.8}     \\
\toprule[1.2pt]
\multicolumn{11}{c}{Pascal-VOC}\\ \hline
ZS3Net& 37.9 & 61.1 & 90.8  & 73.5 &69.3 & \textbf{92.9}   &78.7  & 26.1 &  46.7  &  51.5   \\ 
\textbf{CaGNet}&\textbf{50.9}  &\textbf{63.2}   & \textbf{91.4}   &\textbf{74.6}   & \textbf{69.5}  & 92.7   & \textbf{78.9} &\textbf{40.2}   & \textbf{67.8}     & \textbf{52.3} \\ \bottomrule[1.5pt]
\end{tabular}
\caption{Zero-shot segmentation performances on Pascal-Context and Pascal-VOC datasets in the setting of ZS3Net. The best results are denoted in boldface.}
\label{exp1}
}
\end{table*}

\section{Comparison in the setting of ZS3Net}

To further verify the effectiveness of our proposed method, we also evaluate our method in the setting of ZS3Net~\cite{bucher2019zero} (\emph{i.e.}, backbone, semantic word embedding method, seen/unseen splits, evaluation metrics). Note that we only follow the setting of ZS3Net~\cite{bucher2019zero} in this section, while in other sections we still follow SPNet~\cite{xian2019semantic}.

Following ZS3Net~\cite{bucher2019zero}, we use word2vec~\cite{mikolov2013distributed} embeddings in length 300 as semantic word embeddings and use deeplabv3+~\cite{deeplabv3} as the backbone. For both evaluation and training, we treat ``background'' as a seen category following ZS3Net. We conduct experiments on Pascal-VOC dataset with 20 categories and Pascal-Context dataset with 59 categories.
For seen/unseen split, we choose one of the splits provided by ZS3Net for each dataset: ``cow, motorbike, airplane, sofa'' as 4 unseen categories on Pascal-VOC dataset, and ``cow, motorbike, sofa, cat, boat, fence, bird, tvmonitor, keyboard, aeroplane'' as 10 unseen categories on Pascal-Context dataset.


The experimental results are shown in Table~\ref{exp1} and the results of ZS3Net are directly copied from their paper. Our method achieves comparable or better results on seen categories. More importantly, our method significantly improves the results on unseen categories.
For overall hIoU, our method achieves the improvement of $13.0$ and $4.9$ on Pascal-VOC and Pascal-Context respectively. This indicates that our method still beats ZS3Net in their setting with dramatic improvements. Another observation is that our method has much larger performance gain on Pascal-VOC than Pascal-Context, which may be due to the difficulty in segmenting more unseen categories.

\begin{table}
\centering
\resizebox{\columnwidth}!{
\begin{tabular}{c|cccc}
\toprule[1.5pt]
  &\textbf{hIoU} &mIoU&S-mIoU&U-mIoU    \\ \hline \hline
w/o residual& 0.3862 & 0.6480  & 0.7815 & 0.2564\\
Paralle& 0.3821 &0.6509   & 0.7832 &0.2527 \\ \hline
CaGNet& \textbf{0.3972} & \textbf{0.6545} & \textbf{0.7840}&\textbf{0.2659} \\ \bottomrule[1.5pt]
\end{tabular}}
\caption{Ablation studies of special cases of the contextual module on Pascal-VOC.}
\label{specialcase}
\end{table}

\section{More Ablation Studies on Our Contextual Module}

In this section, we add two more special cases ``w/o residual" and ``Parallel" to supplement Table 3 of Section 4.4 in the main paper, as part of the ablation studies on different variants of our Contextual Module.

In the special case ``w/o residual",  our Contextual Module ($CM$) outputs the contextual latent code without being linked back to the segmentation network, so that residual attention is not applied to feature map $\textbf{F}_n$ to obtain enhanced feature map $\textbf{X}_n$. 
In this case, the contextual latent code is obtained in the same way, while the target of feature reconstruction becomes $\textbf{F}_n$ instead of $\textbf{X}_n$.
We also replace $\textbf{X}_n$ with $\textbf{F}_n$ in all loss functions. The results are shown in the first row of Table~\ref{specialcase}. We can observe that linking our $CM$ to the segmentation network improves the performances on all metrics.

In the special case ``Parallel", we change the way of arranging three dilated conv layers in $CM$ from serial to parallel.
That is, we parallelly put three dilated conv layers (same parameters as those in original $CM$ respectively) after the input feature map $\textbf{F}_n$ and obtain context maps of different receptive fields. The receptive fields of three dilated convs are $3\times3$, $5\times 5$, and $13\times 13$ on $\textbf{F}_n$ respectively, which are equal to or smaller than those ($3\times3$, $7\times 7$, and $17\times 17$) in the serial mode. The experiment results of the second row in Table~\ref{specialcase} indicates that the serial mode in the main paper is more superior than the parallel mode, probably due to the larger receptive fields of the obtained context maps.

\section{Hyper-parameter Analyses}

\begin{figure}[]
\centering
\includegraphics[width=\linewidth]{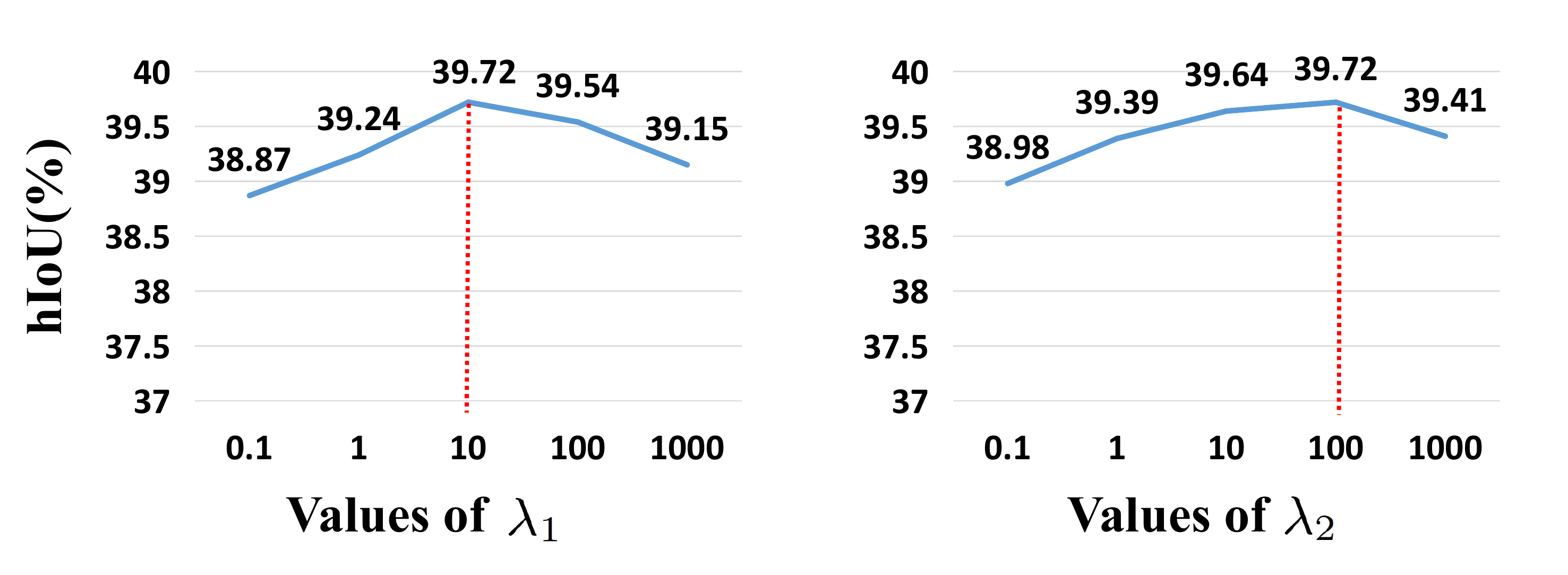}
\caption{The \textbf{effects} of varying the values of $\lambda_{1},\lambda_{2}$ on Pascal-VOC. The dashed lines denote the default values used in our paper.}\label{alpha}
\end{figure}

By taking Pascal-VOC dataset as an example, we investigate the impact of hyper-parameters $\lambda_{1}, \lambda_{2}$ in our method (depicted in Section 3.4 in the main paper). We vary $\lambda_{1}$ (\emph{resp.}, $\lambda_{2}$) within the range [0.1,1000] and report hIoU (\%) results of our method in Figure~\ref{alpha}. We observe that $\lambda_1$ has larger impact and the performance drops sharply when $\lambda_1$ is very small, which proves the necessity of feature reconstruction. Our method is robust when setting $\lambda_1$ (\emph{resp.}, $\lambda_2$)  in a reasonable range [10, 100] (\emph{resp.}, [1, 1000]).

\begin{figure}
\centering
\includegraphics[width=\linewidth]{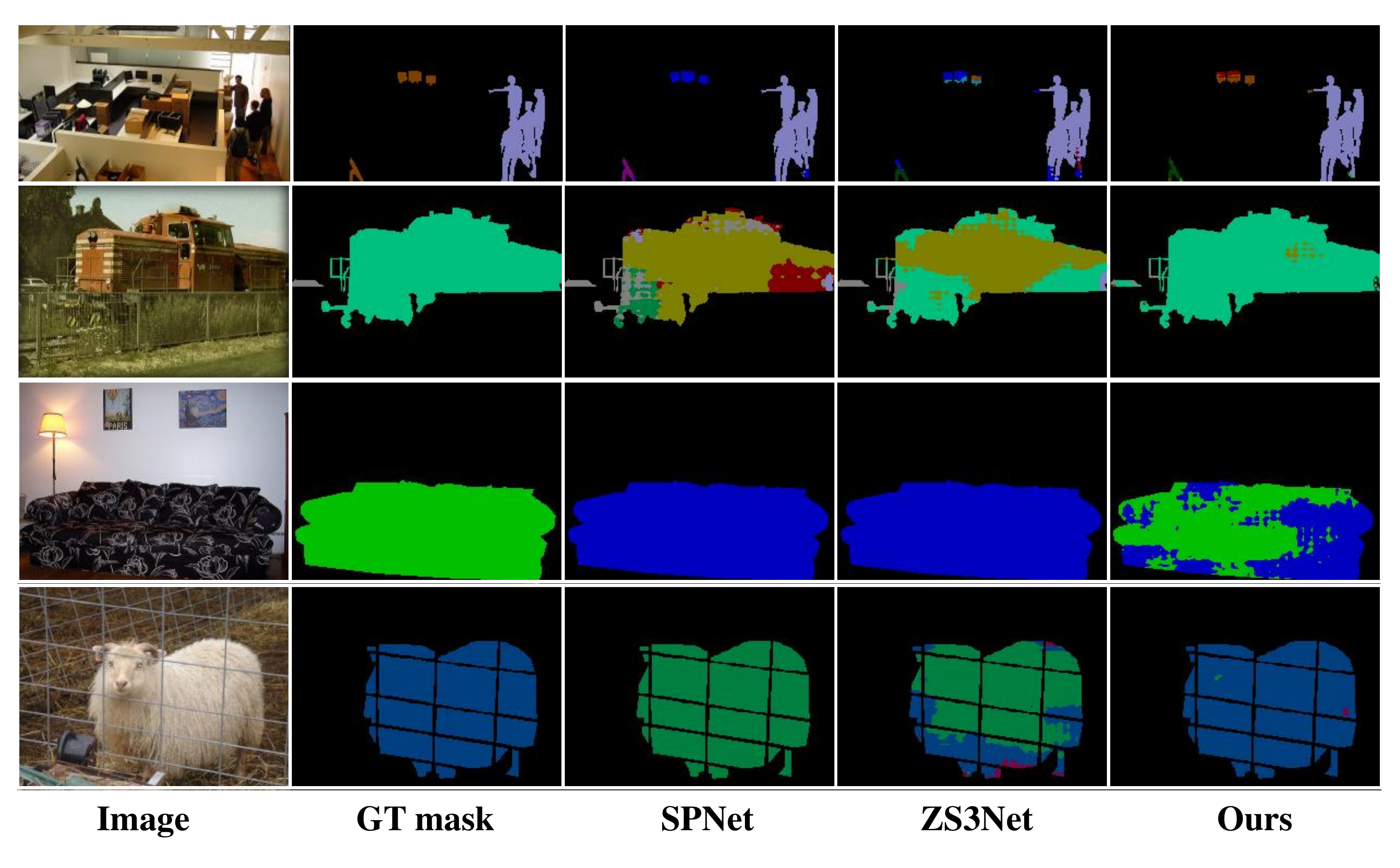}
\caption{Visualization of segmentation results for different methods on Pascal-VOC dataset. GT mask is the ground-truth segmentation mask.}
\label{vis1}
\end{figure}

\begin{figure}
\centering
\includegraphics[width=\linewidth]{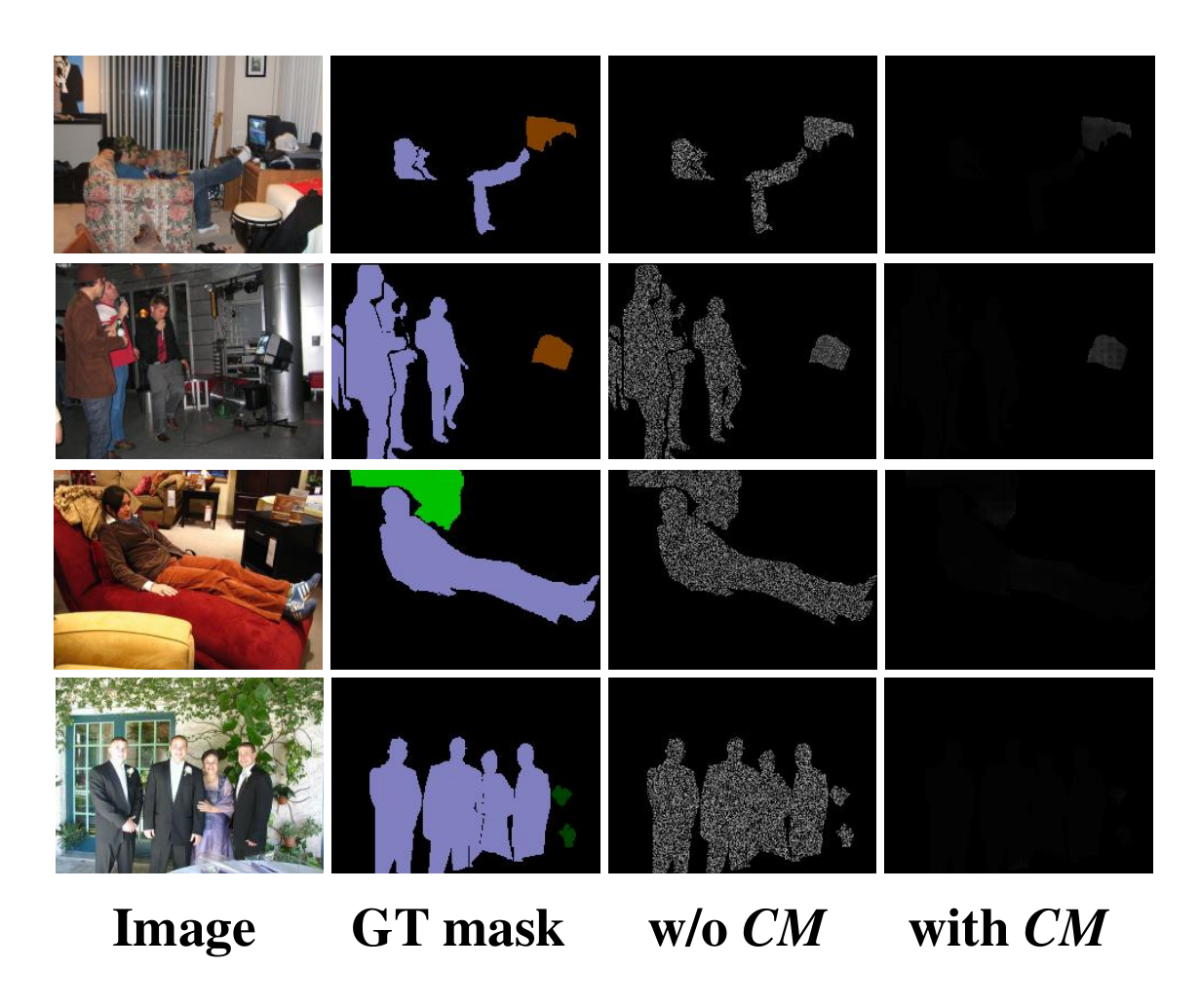}
\caption{Visualization of the effectiveness of Contextual Module (CM) in feature generation on test images on Pascal-VOC dataset. In the second column, GT mask is the ground-truth segmentation mask. In the third and fourth columns, we show the reconstruction loss maps calculated based on the generated feature maps and real feature maps (the darker, the better).}
\label{vis2}
\end{figure}

\section{More visualizations of segmentation results}

In this section, we show more visualizations of segmentation results for different methods in Figure~\ref{vis1}, supplementing the visualizations in Figure 5 of Section 4.6 in the main paper.

As shown in Figure~\ref{vis1}, our method beats others when segmenting unseen objects like ``tv", ``train'', ``sofa'', and ``sheep'', which further proves the advantage of our method. For example, in the first and third row, SPNet and ZS3Net misclassify ``tv'' and ``sofa'' as ``table'', but our method segments them successfully. We can also observe that ``train'' in the second row is hard to segment by SPNet and ZS3Net. This is probably because the word ``train'' contains several distinct meanings and only one of them represents the typical unseen category in the dataset. Therefore, the semantic word embedding of ``train'' is not accurate enough for the model to segment objects of this category precisely. However, our method can still recognize and segment it. In the fourth row, ``sheep'' is also recognized by our method, while ZS3Net and SPNet classify it as ``cow''.

\section{More visualizations of feature generation}

We show more visualizations of feature generation in Figure~\ref{vis2}, supplementing the visualizations in Figure 6 of Section 4.6 in the main paper. By taking test images of Pascal-VOC dataset as examples, we show the reconstruction loss maps calculated based on the generated feature maps and their according real feature maps, in which smaller loss (darker region) implies better generation quality. We compare the reconstruction loss maps obtained by using Contextual Module ($CM$) or without $CM$. It can be observed from Figure~\ref{vis2} that our $CM$ not only facilitates generating better features for seen categories (\emph{e.g.}, ``person''), but also for unseen categories (\emph{e.g.}, ``tv'' in brown in the first two rows, ``potted plant'' in dark green in the fourth row).

\bibliographystyle{ACM-Reference-Format}
\bibliography{ijcai20}